\documentclass{article}

\usepackage[main, final]{neurips_2026}

\usepackage[utf8]{inputenc} 
\usepackage[T1]{fontenc}    
\usepackage{hyperref}       
\usepackage{url}            
\usepackage{booktabs}       
\usepackage{amsfonts}       
\usepackage{nicefrac}       
\usepackage{microtype}      
\usepackage[table]{xcolor}  
\usepackage{amsmath}
\usepackage{fontawesome5}
\usepackage{graphicx}
\usepackage{cleveref}
\usepackage{caption}
\usepackage{multirow}
\usepackage{placeins} 
\usepackage{pifont}
\crefname{figure}{Fig.}{Figs.}
\Crefname{figure}{Fig.}{Figs.}
\crefname{table}{Tab.}{Tabs.}
\Crefname{table}{Tab.}{Tabs.}
\crefname{section}{Sec.}{Secs.}
\crefname{equation}{Eq.}{Eqs.}

\usepackage{booktabs}
\usepackage{multirow}

\definecolor{BestBG}{HTML}{BFE7CC}
\definecolor{SecondBG}{HTML}{B9E3F0}
\definecolor{ThirdBG}{HTML}{CCD8F6}

\newcommand{\rankfirst}[1]{\cellcolor{BestBG}\textbf{#1}}
\newcommand{\ranksecond}[1]{\cellcolor{SecondBG}#1}
\newcommand{\rankthird}[1]{\cellcolor{ThirdBG}#1}

\title{MCSplat: Multi-View Photometric and Geometric Consistent Feed-Forward Gaussian Splatting for Driving Scenes}

\author{
Yifei Wang\textsuperscript{1,2}\thanks{
  Work done during an internship at Zaofu Intelligent Technology Co., Ltd.
},
Wei Tian\textsuperscript{3,\faEnvelope[regular]},
Wenhua Bao\textsuperscript{2},
Jia Zhou\textsuperscript{3}, \\
{\bfseries
Hanshi Wang\textsuperscript{4},
Fudong Ge\textsuperscript{4},
Jianhua Wu\textsuperscript{3},
Shuai Yuan\textsuperscript{5},
Zhipeng Zhang\textsuperscript{5},
}\\
{\bfseries
Xingliang Liu\textsuperscript{2,$\dagger$},
Xianming Zeng\textsuperscript{2,$\dagger$},
Jianyun Xu\textsuperscript{2,$\dagger$},
Bingzhao Gao\textsuperscript{1,3,\faEnvelope[regular]}
}
\\[0.1em]
\textsuperscript{1}Shanghai Research Institute for Intelligent Autonomous Systems, Tongji University\\
\textsuperscript{2}Zaofu Intelligent Technology Co., Ltd.\\
\textsuperscript{3}College of Automotive and Energy Engineering, Tongji University\\
\textsuperscript{4}Institute of Automation, Chinese Academy of Sciences, \textsuperscript{5}SAI, Shanghai Jiao Tong University\\
{\normalfont\small
\url{https://yfwang0824.github.io/mcsplat-project/}
}
}

\begin{document}
\maketitle
\begingroup
\renewcommand{\thefootnote}{\ensuremath{\dagger}}
\footnotetext{
  Project leaders of Zaofu Intelligent Technology Co., Ltd.
}

\renewcommand{\thefootnote}{\faEnvelope[regular]}
\footnotetext{
  Corresponding authors:
  Wei Tian (\texttt{tian\_wei@tongji.edu.cn})
  and Bingzhao Gao (\texttt{gaobz@tongji.edu.cn}).
}
\endgroup

\vspace{-2.0em}  

\begingroup
\centering
\includegraphics[width=\textwidth]{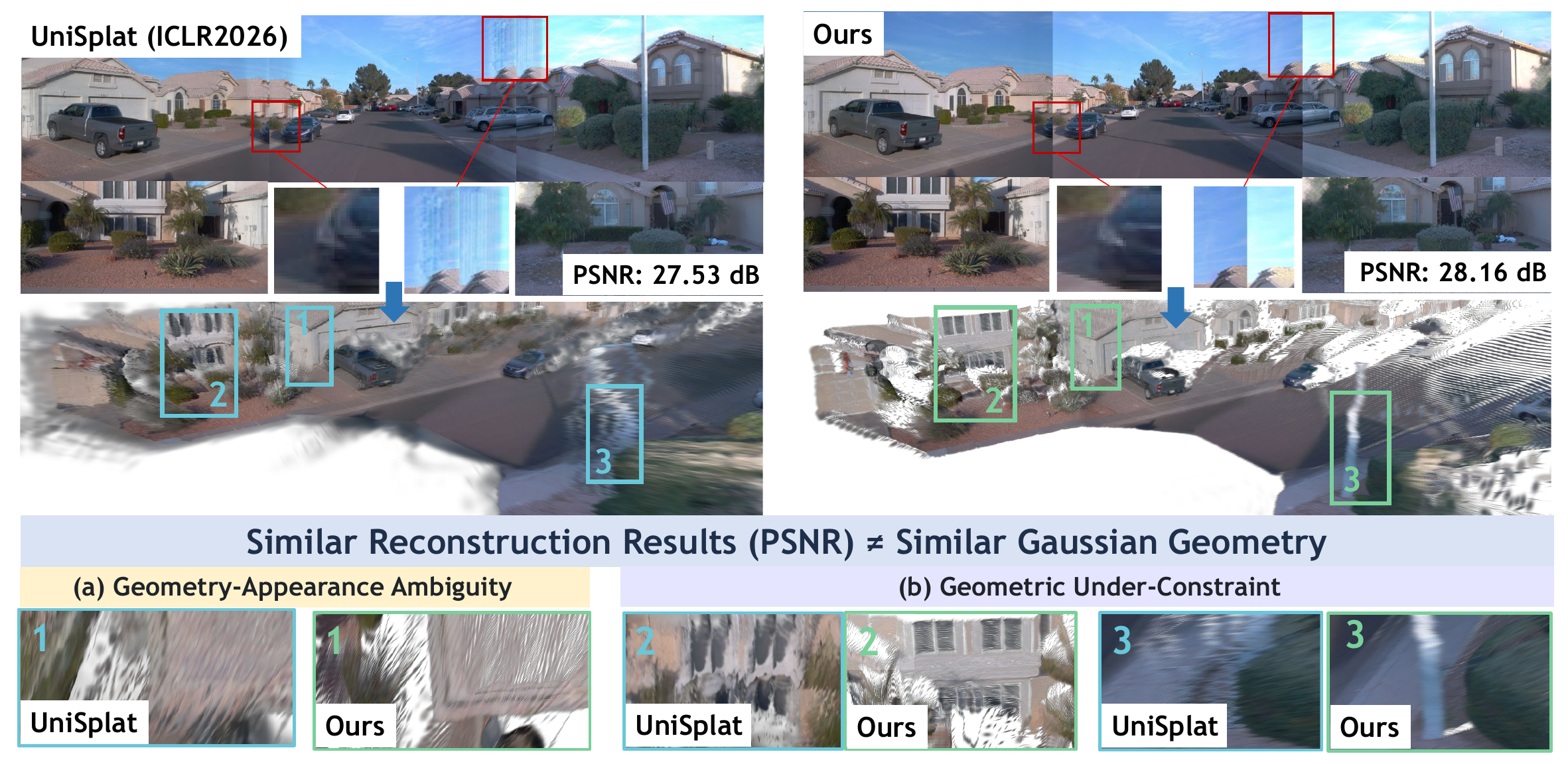}

\captionsetup{
  justification=justified,
  singlelinecheck=false
}
\captionof{figure}{
As shown by the comparison with
UniSplat~\citep{shi2026unisplat},
comparable PSNR can still correspond to fundamentally
different Gaussian representations.
MCSplat addresses
\textbf{(a) cross-view geometry-appearance ambiguity}
and
\textbf{(b) local geometric supervision unreliability}
by combining multi-view photometric canonicalization
with uncertainty-aware geometric self-supervision,
producing cleaner Gaussian geometry and more
consistent structures in cross-camera overlap regions.
}
\label{fig:teaser}
\par
\endgroup


\begin{abstract}
Driving-scene reconstruction via feed-forward Gaussian splatting
from sparse multi-view images remains challenging due to two factors.
First, cross-camera photometric inconsistency, caused by variations
in exposure, ISP, and illumination, can be incorrectly absorbed
into Gaussian geometry. Second, unreliable local geometric
supervision arises from normal-based constraints near depth
discontinuities.
To address these issues, we present \textbf{MCSplat},
a multi-view consistent feed-forward Gaussian splatting framework
for driving scenes. MCSplat constructs a canonical appearance
space from depth-guided multi-camera correspondences to decouple
geometry learning from camera-specific appearance.
A multi-scale Bilateral Grid Transformer restores raw camera
appearance from canonical renderings, enabling photometric
supervision without contaminating Gaussian geometry.
To improve local geometric reliability, we estimate
Fisher-information-based uncertainty from photometric sensitivity
and use it to adaptively weight surface-oriented self-supervision.
Extensive experiments demonstrate
improved reconstruction quality and generalization across
driving datasets. Geometry-oriented evaluations further reveal
enhanced cross-view depth consistency and local surface-normal
agreement, while qualitative results show fewer geometric
artifacts and more coherent structures across overlapping views.
\end{abstract}

\section{Introduction}

Neural rendering methods such as
3D Gaussian Splatting (3DGS)~\citep{kerbl20233d} have enabled high-quality 3D reconstruction and novel view synthesis, making urban driving scene reconstruction as a cornerstone of closed-loop autonomous driving simulation~\citep{yang2023unisim} and long-horizon planning~\citep{gao2025rad}. Since per-scene optimization remains computationally expensive, recent generalizable splatting frameworks~\citep{chen2025dggt,shi2026unisplat,yu2026streetforward} directly predict Gaussian primitives from multi-frame images. Leveraging Geometry Foundation Models (GFMs)~\citep{wang2025vggt,wang2026pi,keetha2025mapanything,lin2025depth}, they use dense depth predictions as priors for Gaussian means, along with a Gaussian head to predict  rotation, scale, opacity, color, and motion. Despite improved efficiency and generalization, they still inherit two key challenges from 3DGS optimization:

\noindent\textbf{Cross-view geometry-appearance ambiguity:}
Ideally, multi-view supervision should attribute cross-view image differences to the underlying 3D structure rather than camera-dependent photometric effects. In real-world driving scenarios, however, photometric consistency is difficult to guarantee across views and time due to camera-dependent exposure, color response, automatic image signal processing (ISP), and changing illumination. As illustrated in \cref{fig:teaser} (a), inconsistent appearance in cross-camera overlap regions can cause the same structure, such as the roof, to be explained as separate surfaces, producing floating Gaussians and degraded geometry~\citep{martin2021nerf}.

\noindent\textbf{
Local geometric supervision unreliability:
}
Driving scenes provide dense observations along the ego trajectory but sparse off-trajectory views. Even with GFM-predicted depths as Gaussian means, rotation, scale, opacity, and color remain weakly constrained by RGB and depth losses, allowing multiple Gaussian configurations to fit training views while differing in off-trajectory renderings. As illustrated in \cref{fig:teaser} (b), this can lead to geometrically incorrect yet view-consistent Gaussian representations, making cross-lane novel view synthesis challenging. Geometric self-supervision can alleviate this ambiguity by aligning Gaussian orientations with local surface normals. However, surface normals are typically derived from local finite differences of depth or Gaussian centers, which become unreliable near depth discontinuities and may introduce inaccurate supervision~\citep{hosseinzadeh2025g3splat}.

Existing methods attempt to mitigate the geometry-appearance ambiguity using appearance embeddings, global color transforms, explicit ISP parameters, or bilateral-grid-based local color transforms~\citep{martin2021nerf,kulhanek2024wildgaussians,cui2025luminance,deutsch2026ppisp,wang2024bilateral,wang2025unifying}. However, most of these methods rely on per-scene optimization, which incurs high computational cost and limits cross-scene generalization. Recent feed-forward bilateral-grid methods reduce the need for scene-specific optimization by assuming sufficient overlap between reference and target views~\citep{shin2026chroma}, making them less suitable for surround-view driving systems where inter-camera overlap is inherently limited.

Beyond photometric variation, the under-constrained nature of geometric reconstruction is typically addressed by introducing additional supervisory signals. Existing methods employ semantic, flow, or normal priors to constrain geometry~\citep{peng2025desire,zhou2024hugs,wang2025drivesplat}. However, these priors rely on strong teacher models and are costly to scale to feed-forward reconstruction, as they require additional data preparation, storage, and training overhead. Self-supervised geometric constraints reduce the reliance on extra priors but remain unreliable near depth discontinuities or when depth estimates from foundation models are inaccurate~\citep{hosseinzadeh2025g3splat}.

In this work, we propose to enhance the GFM-based feed-forward Gaussian reconstruction for driving scenes by combining geometry--appearance decoupling with uncertainty-aware geometric self-supervision. To mitigate the geometry--appearance ambiguity, we provide appearance-consistent inputs to the feed-forward Gaussian model, thereby reducing the impact of cross-view and cross-time photometric discrepancies on Gaussian prediction. A lightweight transformer predicts multi-scale 3D bilateral grids to map rendered images back to the original GT appearance, enabling photometric supervision while preserving view-specific imaging effects. To handle local geometric supervision unreliability, we use Fisher information with respect to the photometric loss to quantify geometric uncertainty and construct an uncertainty-aware self-supervised geometric loss.

Our contributions are summarized as follows.
\textbf{First,} we propose \textbf{MCSplat}, a GFM-compatible geometry--appearance decoupling framework that builds a canonical appearance space and recovers camera-specific appearance with a multi-scale Bilateral Grid Transformer, reducing cross-view geometry--appearance ambiguity.
\textbf{Second,} we introduce Fisher information-guided geometric self-supervision that estimates geometric uncertainty from photometric sensitivity and adaptively weights surface regularization, alleviating unreliable local supervision.
\textbf{Third,} our method achieves state-of-the-art results on the Waymo benchmark, including an appearance-challenging subset, while producing more geometrically consistent novel-view reconstructions.

\section{Related Works}
\label{gen_inst}

\paragraph{Neural Scene Reconstruction for Driving Scenes.}NeRF~\citep{mildenhall2021nerf} enables high-quality novel view synthesis but is inefficient for rendering,
while 3DGS~\citep{kerbl20233d} uses explicit learnable Gaussians for real-time rendering, making it suitable for dynamic driving scenes. Many driving-scene methods adopt per-scene optimization with priors, including LiDAR~\citep{zhou2024drivinggaussian}, semantics~\citep{yan2024street}, optical flow~\citep{zhou2024hugs}, human models~\citep{chen2024omnire}, scene flow~\citep{xu2025ad}, and multi-scale voxels~\citep{wang2025drivesplat}. 
However, scene-specific retraining limits scalability and cross-scene generalization. 
Feed-forward methods instead directly regress Gaussian representations from images. STORM~\citep{yang2024storm} predict pixel-aligned Gaussian motion.
DrivingForward~\citep{tian2025drivingforward} and DrivingScene~\citep{hou2025drivingscene} infer Gaussian primitives from surround-view inputs. 
Recent GFM-based methods, including DGGT~\citep{chen2025dggt}, StreetForward~\citep{yu2026streetforward}, UniSplat~\citep{shi2026unisplat}, and ReconDrive~\citep{yu2026recondrive}, leverage GFMs~\citep{wang2025vggt,wang2026pi,keetha2025mapanything,lin2025depth} to constrain Gaussian positions and decode Gaussian attributes. Despite improved efficiency and generalization, these methods remain vulnerable to geometry--appearance ambiguity under cross-camera photometric inconsistency.
Additionally, they are prone to geometrically under-constrained reconstruction due to unreliable geometric supervision.

\paragraph{Appearance Harmonization in Neural Scene Representations.}
Cross-camera photometric inconsistency is inevitable in real world and induces geometry--appearance ambiguity. NeRF in the Wild~\citep{martin2021nerf} and WildGaussians~\citep{kulhanek2024wildgaussians} model imaging variation with appearance branches or observation-dependent embeddings. STORM~\citep{yang2025storm} learns coarse global affine transforms for feed-forward reconstruction.
PPISP~\citep{deutsch2026ppisp} explicitly parameterizes ISP factors such as exposure, vignetting, and camera response functions. However, these methods mainly rely on global or per-scene optimization, making them less suitable for handling local and cross-camera discrepancies.
Thus, refined photometric operation has been employed in recent methods. 
Bilateral-grid~\citep{chen2007real} provides spatially varying photometric transforms. BilaRF~\citep{wang2024bilateral} uses bilateral-space affine transforms for pixel-wise color adjustment. 
CHROMA~\citep{shin2026chroma} explores feed-forward bilateral-grid prediction for multi-view harmonization. However, these methods require substantial overlap between reference and target views, limiting their applicability to surround-view driving scenes with small inter-camera overlap.

\paragraph{Uncertainty Quantification in Neural Scene Reconstruction.}
Uncertainty estimation measures reconstruction reliability in challenging regions such as occlusion boundaries, weak textures, and dynamic objects~\citep{jiang2023fisherrf,kim20244d,wang2026uncertainty}. Fisher information~\citep{fisher1922mathematical} provides a parameter-sensitivity-based uncertainty measure and has been used for active view selection, pruning, and robust reconstruction.
FisherRF~\citep{jiang2023fisherrf} uses expected information gain for active view selection.
PUP-3DGS~\citep{hanson2025pup} adopts sensitivity scores for Gaussian pruning.
OUGS~\citep{li2025ougs} propagates Gaussian-parameter uncertainty to pixel space for object-aware view selection. 
However, uncertainty has primarily been explored in the context of per-scene optimization, with its role in feed-forward reconstruction remaining limited. In contrast, we compute geometry uncertainty from the photometric loss within a feed-forward framework, and leverage it for adaptive geometric self-supervision.

\section{Methodology}
\label{headings}

\begin{figure}[t]
\centering
\vspace{-1.0em}
\includegraphics[width=\linewidth]{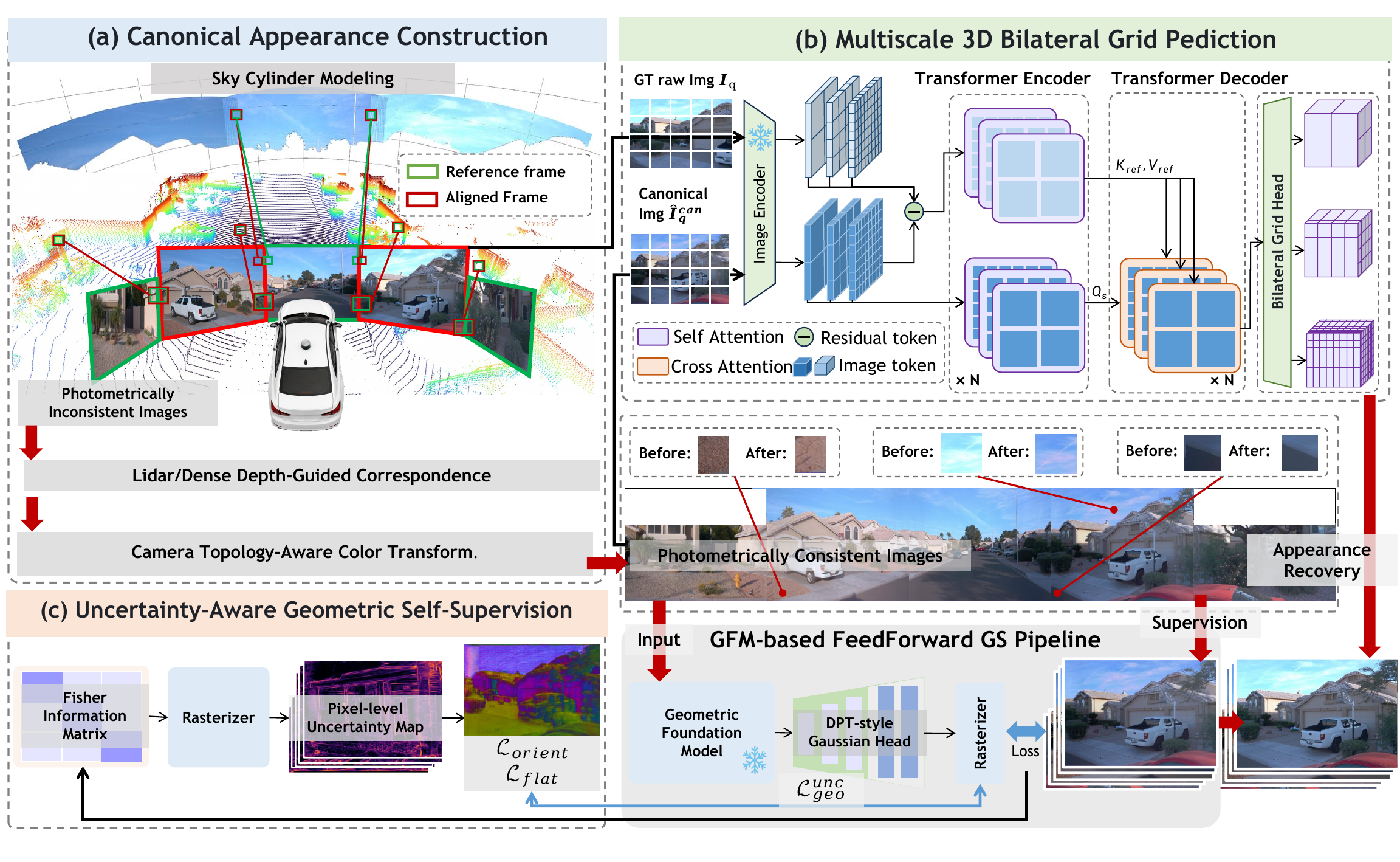}
\caption{
Overview of MCSplat. 
(a) Depth-guided canonical appearance construction aligns inconsistent multi-view images.
(b) A multi-scale bilateral-grid transformer restores camera-specific appearance from canonical renderings. 
(c) Fisher information-based uncertainty guides geometric self-supervision in under-constrained regions.
}
\vspace{-1.0em}
\label{fig:pipeline}
\end{figure}

We propose \textbf{MCSplat}, a framework for feed-forward reconstruction of multi-view driving scenes 
by integrating geometry--appearance decoupling with uncertainty-aware geometric self-supervision, shown in \cref{fig:pipeline}. 
Its goal is to predict reliable 3D Gaussian primitives from multi-view images, enabling novel-view synthesis for closed-loop autonomous driving simulation.

In our framework, given photometrically inconsistent multi-view images, we first construct a \emph{canonical appearance space} using depth-guided correspondences, and learn Gaussian primitives within this normalized domain. To recover the original, camera-specific appearance, a lightweight transformer predicts multi-scale 3D bilateral grids and applies them to rendered images in a coarse-to-fine manner. We further estimate pixel-wise geometric uncertainty based on photometric-loss sensitivity, and use it to adaptively weight geometric self-supervision in weakly constrained regions. 

\definecolor{mygreen}{RGB}{0,150,0}
\definecolor{myred}{RGB}{175,0,0}
\subsection{Robust Canonical Appearance Construction}
\label{sec:robust_canonical_appearance}
To reduce geometry--appearance ambiguity at the input stage, we construct a canonical appearance space using depth-guided correspondences. Given a multi-view camera set \(\mathcal{V}\), reference cameras \(\mathcal{V}_{\mathrm{ref}}\subset\mathcal{V}\) are selected according to rig topology and inter-camera overlap. Reference views define the canonical appearance space with identity transforms, while non-reference views are aligned to adjacent reference anchors. For example, in the five-camera Waymo benchmark layout shown in \cref{fig:pipeline} (a), we use
\(
\mathcal{V}_{\mathrm{ref}}=
\{\textcolor{mygreen}{\mathrm{Front}},
\textcolor{mygreen}{\mathrm{SideLeft}},
\textcolor{mygreen}{\mathrm{SideRight}}\}
\).
Consequently, the two remaining cameras, \textcolor{myred}{FrontLeft} and \textcolor{myred}{FrontRight} are each anchored by two neighboring references.

For each 
pixel \({p}_i\) in a reference camera view \(r\in\mathcal{V}_{\mathrm{ref}}\)
with a 
valid depth $D_r({p}_i)$, 
we back-project it to 3D coordinates 
and then reproject it into a new pixel $\tilde{{p}}_i$ in
a target camera view \(t\), where \(t\) is selected from the non-reference camera set \(\mathcal{V}\setminus\mathcal{V}_{\mathrm{ref}}\).
For sky regions without reliable depth, we establish correspondences by intersecting each reference-camera ray with a distant ego-centered sky cylinder. The resulting 3D intersection point is then projected into the target camera, yielding a sky-region matched pixel.

Given correspondences between \(r\) and \(t\), we estimate a per-channel diagonal affine transform \(\mathbf{A}_{t\rightarrow r}\) (with parameters $a_c, b_c$ for image channel $I^c$) from target to reference appearance:
\begin{equation}
I_r^c({p}_i)
\approx
a_c I_t^c(\tilde{{p}}_i)+b_c,
\qquad
c\in\{R,G,B\},
\end{equation}
by solving the optimization problem with hyperparameters $\lambda_a, \lambda_b$:
\begin{equation}
\min_{\{a_c,b_c\}}
\sum_i \rho\!\left(a_c I_t^c(\tilde{{p}}_i)+b_c-I_r^c({p}_i)\right)
+\lambda_a\sum_c(a_c-1)^2
+\lambda_b\sum_c b_c^2 .
\end{equation}
Here, \(\rho(\cdot)\) is a robust penalty, and the identity prior makes the transform fall back to no correction when correspondences are sparse or unreliable. We additionally use a shared-gain fallback for chromatically degenerate samples, a few IRLS reweighting steps, and residual-based inlier refitting when enough inliers are available.

For target cameras with two adjacent references, to avoid biased single-reference alignment and discontinuous hard switching, we interpolate target-to-reference transforms along the horizontal image axis. 
Given a horizontal image coordinate \(u\in[0,W-1]\) and the ratio \(\alpha(u)=u/(W-1)\), for a target camera \(t\) with adjacent references \(r_1\) and \(r_2\), the spatially varying affine field is
\begin{equation}
\mathbf{A}_{t}(u)
=
(1-\alpha(u))\,\mathbf{A}_{t\rightarrow r_1}
+
\alpha(u)\,\mathbf{A}_{t\rightarrow r_2}.
\end{equation}
Sky and non-sky regions  share the same interpolation form but use separately estimated affine transforms, accounting for their distinct color statistics.

\subsection{Multi-Scale Bilateral Grid Transformer for Appearance Recovery}
\label{sec:bilateral_grid_head}


Given the canonical render \(\hat{I}_q^{\mathrm{can}}\), we recover the original camera-specific appearance \(I_q\) without re-entangling photometric variation with geometry. For each query view, we predict multi-scale bilateral transforms and apply them to \(\hat{I}_q^{\mathrm{can}}\) in a coarse-to-fine manner, enabling compact modeling of global exposure shifts, color biases, and spatially varying local corrections.



\paragraph{3D bilateral-grid parameterization.}
At a given scale \(m\), the appearance transform is represented by a 3D bilateral grid
\begin{equation}
\mathbf{B}_q^{(m)} \in \mathbb{R}^{12 \times L_m \times H_m \times W_m},
\end{equation}
where \(H_m,W_m\) are spatial resolutions and \(L_m\) is the guidance resolution. 
Each grid cell stores a flattened \(3\times4\) affine color transform. 
For a pixel \(p\), we slice the grid using its normalized coordinates \((x_p^{(m)},y_p^{(m)})\) and grayscale guidance
\(\gamma_p=\mathrm{gray}(\hat{I}_q^{\mathrm{can}}(p))\):
\begin{equation}
\boldsymbol{\theta}_{q,p}^{(m)}
=
\mathcal{S}\!\left(\mathbf{B}_q^{(m)},x_p^{(m)},y_p^{(m)},\gamma_p\right),
\end{equation}
where \(\boldsymbol{\theta}_{q,p}^{(m)}\) is a \(3\times4\) affine color transform and reshaped into an affine pair
\((\mathbf{A}_{q,p}^{(m)},\mathbf{b}_{q,p}^{(m)})\). The multi-scale transforms are composed in homogeneous coordinates as
\begin{equation}
\mathbf{T}_{q,p}
=
\mathbf{T}_{q,p}^{(M)}\cdots\mathbf{T}_{q,p}^{(1)},
\qquad
\mathbf{T}_{q,p}^{(m)}
=
\begin{bmatrix}
\mathbf{A}_{q,p}^{(m)} & \mathbf{b}_{q,p}^{(m)}\\
\mathbf{0}^{\top} & 1
\end{bmatrix},
\qquad
m \in \{1, \cdots, M\},
\end{equation}
and the recovered image is obtained by
\begin{equation}
\label{eq:trans}
[\hat{I}_q(p),1]^\top
=
\mathbf{T}_{q,p}
[\hat{I}_q^{\mathrm{can}}(p),1]^\top .
\end{equation}
Here, grids from coarse scales capture global shifts while fine grids refine local photometric variations.

\paragraph{Residual-token bilateral grid transformer decoder.}
As illustrated in \cref{fig:pipeline} (b), we predict bilateral grids using a lightweight transformer decoder built on multi-scale token interactions. For a view \(q\), a frozen image encoder extracts patch tokens
\(
P_q^{\mathrm{can}},P_q^{\mathrm{raw}}\in\mathbb{R}^{P\times D}
\)
from the canonicalized and raw images, respectively. At each scale \(m\), we pool them to the target grid resolution:
\begin{equation}
Q_q^{(m)}
=
\operatorname{Pool}\!\left(P_q^{\mathrm{can}};H_m,W_m\right),
\qquad
R_q^{(m)}
=
\operatorname{Pool}\!\left(P_q^{\mathrm{raw}}-P_q^{\mathrm{can}};H_m,W_m\right).
\end{equation}
Here, \(Q_q^{(m)}\) serves as canonical-domain query tokens, while \(R_q^{(m)}\) provides residual key--value tokens. Compared with raw tokens, residual tokens better isolate the photometric transform to be recovered and reduce content redundancy in cross-attention.


For each scale, with the learned 2D positional embedding \(\mathbf{E}_m\), self-attention \(\operatorname{SA}_m\) captures spatial dependencies in the canonical token grid, followed by cross-attention \(\operatorname{CA}_m\) that injects residual appearance cues:
\begin{equation}
Q_q^{(m,1)}=\operatorname{SA}_m\!\left(Q_q^{(m)}+\mathbf{E}_m\right),
\qquad
Q_q^{(m,2)}=\operatorname{CA}_m\!\left(Q_q^{(m,1)},R_q^{(m)}+\mathbf{E}_m\right).
\end{equation}
The refined tokens are passed into a feedforward network 
and mapped to bilateral-grid coefficients through a scale-specific head:
\begin{equation}
\mathbf{B}_q^{(m)}
=
\operatorname{Head}_m\!\left(
\operatorname{FFN}_m\!\left(Q_q^{(m,2)}\right)
\right)
\in
\mathbb{R}^{12\times L_m\times H_m\times W_m}.
\end{equation}
This design captures transforms from global shifts to fine local corrections, while residual cross-attention focuses the prediction on raw-to-canonical appearance differences.

\subsection{Uncertainty-Aware Geometric Self-Supervision}
\label{sec:uncertainty_aware_geo}

\paragraph{Fisher information-based geometry uncertainty.}
For each Gaussian \(g_i\), we estimate its geometric uncertainty by evaluating the sensitivity of the photometric loss to its center \(\boldsymbol{\mu}_i \in \mathbb{R}^3\). Over the current training views \(\mathcal{V}_{\mathrm{cur}}\), we adopt a smooth Charbonnier photometric loss:
\begin{equation}
\mathcal{L}_{\mathrm{photo}}
=
\sum_{q \in \mathcal{V}_{\mathrm{cur}}}
\sum_{p}
\sqrt{
\left(
\hat{I}_q^{\,\text{can}}(p) - I_q^{\,\text{can}}(p)
\right)^2
+
\epsilon^2
}.
\end{equation}
Here, \(p\) indexes valid pixels in view \(q\). \(\epsilon>0\) is a small constant that ensures differentiability and numerical stability.

We estimate a scalar translational Fisher-information score for each Gaussian using the squared photometric-loss gradient norm:
\begin{equation}
F_{i,q}^{xyz}
=
\left\|
\frac{\partial \mathcal{L}_{\mathrm{photo},q}}{\partial \boldsymbol{\mu}_i}
\right\|_2^2,
\qquad
\bar{F}_i^{xyz}
=
\frac{1}{|\mathcal{V}_{\mathrm{cur}}|}
\sum_{q \in \mathcal{V}_{\mathrm{cur}}}
F_{i,q}^{xyz}.
\end{equation}
Intuitively, a large score indicates that small perturbations of the Gaussian center lead to strong photometric changes, implying that its position is well constrained by image evidence. Conversely, a small score indicates weak sensitivity and thus high geometric uncertainty.

\begin{figure}[!t]
\centering
\includegraphics[width=\linewidth]{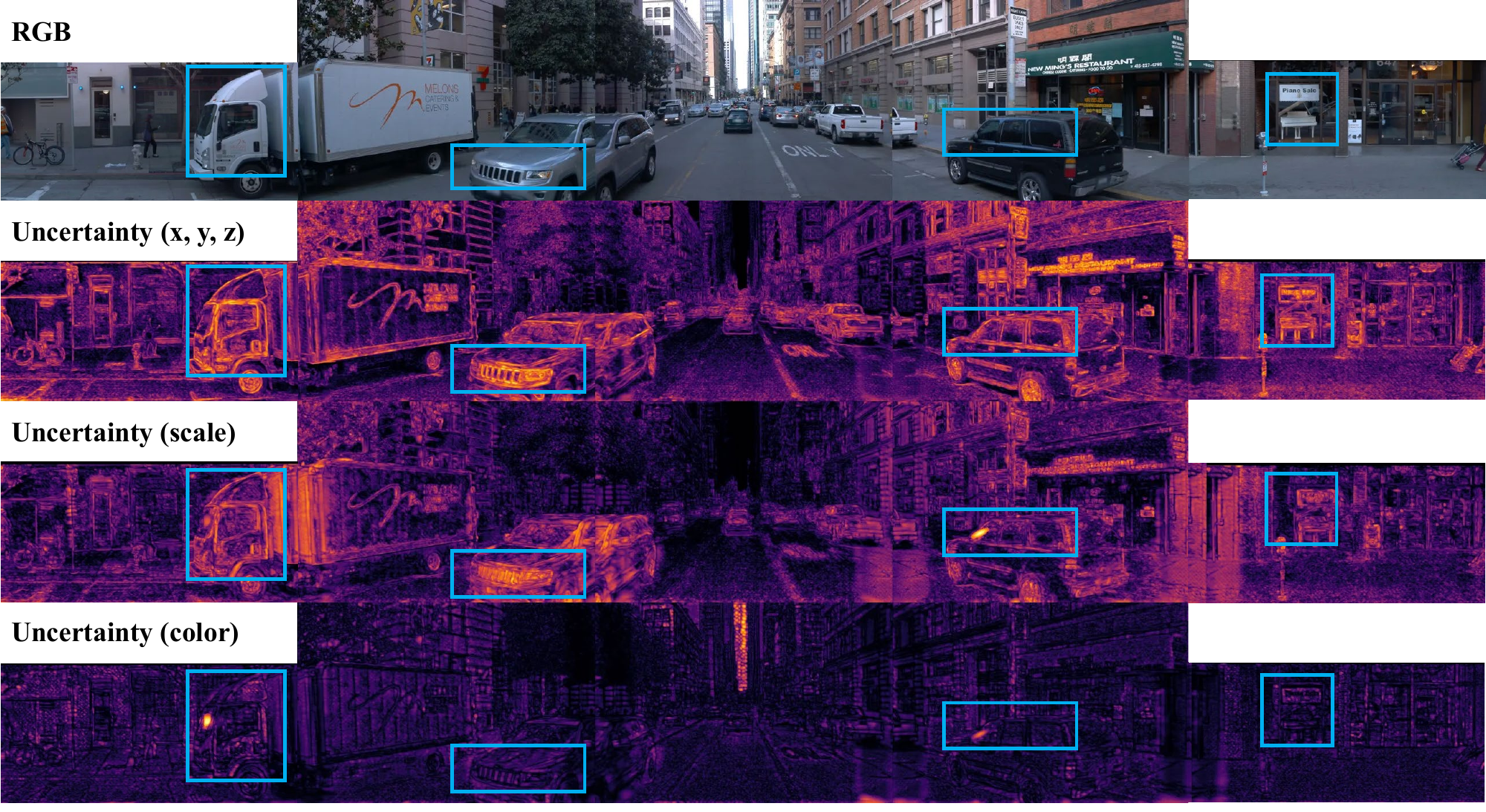}
\caption{
Visualization of photometric-loss uncertainty maps for Gaussian color, scale, and center parameters.
The center-based map captures local structure around the white truck boundary and the jeep grille more distinctly than the scale map and the color map.
}
\label{fig:fisher_vis1}
\end{figure}

\paragraph{Why Gaussian-center Fisher information?}
We use photometric-loss sensitivity to Gaussian centers $(x,y,z)$ as a geometric uncertainty cue, as it more directly reflects positional constraints than color or scale sensitivity. As shown in \cref{fig:fisher_vis1}, the center-based uncertainty map better captures object boundaries and fine structures (e.g., the white truck boundary and jeep grille), whereas color- and scale-based maps are less discriminative. This motivates its use for adaptive geometric supervision.

We normalize the center-based Fisher scores within each mini-batch and convert them into uncertainty weights:
\begin{equation}
\tilde{F}_i^{xyz}
=
\frac{\bar{F}_i^{xyz}}{\max_j \bar{F}_j^{xyz} + \varepsilon},
\qquad
w_i
=
\exp\!\left(-\gamma \tilde{F}_i^{xyz}\right),
\end{equation}
where \(\gamma>0\) controls the sharpness of the weighting. Since \(w_i\) decreases monotonically with Fisher information, Gaussians that are weakly constrained by photometric supervision receive larger geometric regularization weights, while confident Gaussians are regularized less aggressively. In practice, this score is computed online within each training step and detached before weighting the geometric regularizers; implementation details are provided in \cref{app:fisher_reweighting}.

Notably, this uncertainty measure is derived entirely from the model's own photometric sensitivity, without relying on external depth confidence or heuristic visibility priors, making it naturally compatible with feed-forward reconstruction.

\paragraph{Uncertainty-guided surface regularization.}
We apply the uncertainty weights to surface-oriented geometric self-supervision on non-sky Gaussians. Given the predicted Gaussian set, we use the pixel-aligned Gaussian centers \(\{\boldsymbol{\mu}_i\}\) as a 3D point field and estimate local surface normals \(\mathbf{n}_i^{\mathrm{surf}}\) by finite differences over neighboring centers.

For a Gaussian \(g_i\), we define its normal $\mathbf{n}_i^{\mathrm{gs}}$ as the principal axis with the smallest scale $s_{i,k}$:
\begin{equation}
k_i^\star=\arg\min_{k\in\{1,2,3\}}s_{i,k},
\qquad
\mathbf{n}_i^{\mathrm{gs}}
=
\mathbf{R}_i\mathbf{e}_{k_i^\star},
\end{equation}
where 
\(
\mathbf{R}_i
\) 
is the rotation matrix and \(\mathbf{e}_{k_i^\star}\) is the canonical basis vector of the thinnest axis. We then define the uncertainty-weighted orientation loss:
\begin{equation}
\ell_i^{\mathrm{orient}}
=
1-
\left|
\left\langle
\mathbf{n}_i^{\mathrm{surf}},
\mathbf{n}_i^{\mathrm{gs}}
\right\rangle
\right|,
\qquad
\mathcal{L}_{\mathrm{orient}}
=
\frac{\sum_i w_i\,\ell_i^{\mathrm{orient}}}{\sum_i w_i}.
\end{equation}
The above absolute inner product removes normal sign ambiguity and encourages the thinnest Gaussian axis to align with the local surface normal $\mathbf{n}_i^{\mathrm{surf}}$.

To further encourage surface-like Gaussians, we penalize the thinnest scale $s_{i,k_i^\star}$ in loss term:
\begin{equation}
\ell_i^{\mathrm{flat}}=s_{i,k_i^\star},
\qquad
\mathcal{L}_{\mathrm{flat}}
=
\frac{\sum_i w_i\,\ell_i^{\mathrm{flat}}}{\sum_i w_i}.
\end{equation}
The final uncertainty-aware geometric self-supervision loss is defined as
\begin{equation}
\mathcal{L}_{\mathrm{geo}}^{\mathrm{unc}}
=
\lambda_{\mathrm{orient}}\mathcal{L}_{\mathrm{orient}}
+
\lambda_{\mathrm{flat}}\mathcal{L}_{\mathrm{flat}}.
\end{equation}
with hyperparameter weights $\lambda_{\mathrm{orient}}, \lambda_{\mathrm{flat}}$.

\subsection{Training Objective}
\label{sec:training_objective}

We optimize the model in two stages. \textbf{In Stage 1}, the GFM backbone is frozen, while the sparse 3D convolution module and Gaussian head are
trained in the canonical appearance space. The objective is compactly written as
\begin{equation}
\mathcal{L}_{\mathrm{stage1}}
=
\mathcal{L}_{\mathrm{can}}
+
\mathcal{L}_{\mathrm{aux}},
\end{equation}
where \(\mathcal{L}_{\mathrm{can}}\) supervises canonical rendering with MSE and LPIPS losses, and \(\mathcal{L}_{\mathrm{aux}}\) includes dynamic-region and scale-alignment losses. 
\textbf{In Stage 2}, most of the Gaussian head is frozen except for the Gaussian MLPs, while the bilateral-grid transformer is optimized for appearance recovery. The predicted bilateral grids map the canonical render to the GT image domain as in \cref{eq:trans}. The objective becomes
\begin{equation}
\mathcal{L}_{\mathrm{stage2}}
=
\mathcal{L}_{\mathrm{stage1}}
+
\mathcal{L}_{\mathrm{img}}
+
\lambda_7 \mathcal{L}_{\mathrm{tv}}
+
\mathcal{L}_{\mathrm{geo}}^{\mathrm{unc}},
\end{equation}
where \(\mathcal{L}_{\mathrm{img}}\) supervises appearance recovery, \(\mathcal{L}_{\mathrm{tv}}\) regularizes the predicted bilateral grids. The full objective is provided in \cref{app:training_objective}.

\section{Experiments}
\label{experiment}
\subsection{Experimental Settings}


\paragraph{Datasets and Metrics.}
We primarily evaluate on the Waymo Open Dataset~\citep{sun2020scalability}, which contains 798 training and 202 validation scenes, each lasting about 20 seconds at 10 Hz with 5 vehicle-mounted cameras. Following UniSplat~\citep{shi2026unisplat}, we evaluate input-view reconstruction using images at timestamp \(t\) as targets and novel-view synthesis using images at \(t+1\). We report PSNR, SSIM~\citep{wang2004image}, and LPIPS~\citep{zhang2018unreasonable}. \textbf{We also construct an appearance-challenging subset of 57 scenes} selected by input-only inter-camera appearance discrepancy scores from adjacent camera pairs in matched or overlapping regions. More details and results about the subset are provided in \cref{app:choose subset}. We also assess generalization on nuScenes using the protocol described in \cref{sec:experimental_results}.

\paragraph{Baseline.} UniSplat~\citep{shi2026unisplat} is a GFM-based feed-forward Gaussian reconstruction method for dynamic driving scenes. It constructs a 3D latent scaffold from multi-camera videos, performs spatio-temporal fusion in the scaffold space, and decodes dynamic-aware Gaussians using both point and voxel-based branches. We choose UniSplat as our baseline because, to the best of our knowledge, UniSplat is the state-of-the-art feed-forward method trained and evaluated under the full 5-camera setting.

\paragraph{Implementation Details.}
We follow the baseline architecture, including the GFM backbone, semantic feature extraction, and Gaussian head design. The input and rendering resolution is \(350\times518\). Training is performed in two stages 
as described in \cref{sec:training_objective}. 
The Bilateral Grid Transformer uses a DINOv2~\citep{oquab2023dinov2} ViT-Small encoder for appearance token extraction and predicts multi-scale bilateral grids for appearance recovery. We train the model using AdamW~\citep{loshchilov2017decoupled} on 16 NVIDIA H20 GPUs with a total batch size of 32. Additional details are provided in \cref{app:implementation_details}.

\begin{table}[t]
\centering
\caption{
Quantitative comparison of image quality on the Waymo and nuScenes datasets.
The \colorbox{BestBG}{best},
\colorbox{SecondBG}{second-best}, and
\colorbox{ThirdBG}{third-best} results are highlighted.
$^{*}$: Reported by UniSplat.
$^{3}$: Evaluated on the front 3 views only.
}
\label{tab:waymo_nuscenes}

\setlength{\tabcolsep}{3.5pt}
\resizebox{\linewidth}{!}{%
\begin{tabular}{l c ccc ccc ccc}
\toprule

\multirow{3}{*}{Method} &
\multirow{3}{*}{Views} &
\multicolumn{6}{c}{Waymo} &
\multicolumn{3}{c}{nuScenes} \\
\cmidrule(lr){3-8}
\cmidrule(lr){9-11}

& &
\multicolumn{3}{c}{Reconstruction} &
\multicolumn{3}{c}{Novel View Synthesis} &
\multicolumn{3}{c}{Recon\&NVS} \\
\cmidrule(lr){3-5}
\cmidrule(lr){6-8}
\cmidrule(lr){9-11}

& &
PSNR$\uparrow$ & SSIM$\uparrow$ & LPIPS$\downarrow$ &
PSNR$\uparrow$ & SSIM$\uparrow$ & LPIPS$\downarrow$ &
PSNR$\uparrow$ & SSIM$\uparrow$ & LPIPS$\downarrow$ \\

\midrule


EvolSplat$^{*}$~\citep{miao2025evolsplat}
& Front
& \rankthird{23.35} & \rankthird{0.70} & \rankthird{0.29}
& -- & -- & --
& -- & -- & -- \\

UniSplat$^{*}$~\citep{shi2026unisplat}
& Front
& \ranksecond{28.93} & \ranksecond{0.86} & \ranksecond{0.18}
& \ranksecond{27.34} & \ranksecond{0.80} & \ranksecond{0.20}
& -- & -- & -- \\

\textbf{Ours}
& Front
& \rankfirst{31.05} & \rankfirst{0.89} & \rankfirst{0.12}
& \rankfirst{27.64} & \rankfirst{0.84} & \rankfirst{0.16}
& -- & -- & -- \\

\midrule


PixelSplat$^{*}$~\citep{charatan2024pixelsplat}
& Multi
& -- & -- & --
& -- & -- & --
& 21.51 & 0.616 & 0.372 \\

DriveRecon$^{*3}$~\citep{lu2024drivingrecon}
& Multi
& 23.86 & 0.72 & 0.33
& 17.32 & 0.58 & 0.53
& -- & -- & -- \\

MVSplat$^{*}$~\citep{chen2024mvsplat}
& Multi
& 24.94 & \rankthird{0.80} & \rankthird{0.23}
& 22.04 & 0.68 & 0.34
& 21.61 & 0.658 & 0.295 \\

DepthSplat$^{*}$~\citep{xu2025depthsplat}
& Multi
& \rankthird{25.38} & 0.76 & 0.26
& \rankthird{23.86} & \rankthird{0.70} & \rankthird{0.31}
& -- & -- & -- \\

Omni-Scene$^{*}$~\citep{wei2025omni}
& Multi
& -- & -- & --
& -- & -- & --
& \rankthird{24.27} & \rankthird{0.736} & \rankfirst{0.237} \\

UniSplat$^{*}$~\citep{shi2026unisplat}
& Multi
& \ranksecond{29.58} & \ranksecond{0.86} & \ranksecond{0.17}
& \ranksecond{25.98} & \ranksecond{0.76} & \ranksecond{0.24}
& \ranksecond{25.37} & \ranksecond{0.765} & \ranksecond{0.246} \\

\textbf{Ours}
& Multi
& \rankfirst{30.53} & \rankfirst{0.88} & \rankfirst{0.13}
& \rankfirst{26.35} & \rankfirst{0.76} & \rankfirst{0.22}
& \rankfirst{26.19} & \rankfirst{0.783} & \rankthird{0.248} \\

\bottomrule
\end{tabular}%
}
\vspace{-1.0em}
\end{table}

\subsection{Experimental Results}
\label{sec:experimental_results}
\begin{figure}[t]
\centering
\includegraphics[width=\linewidth]
{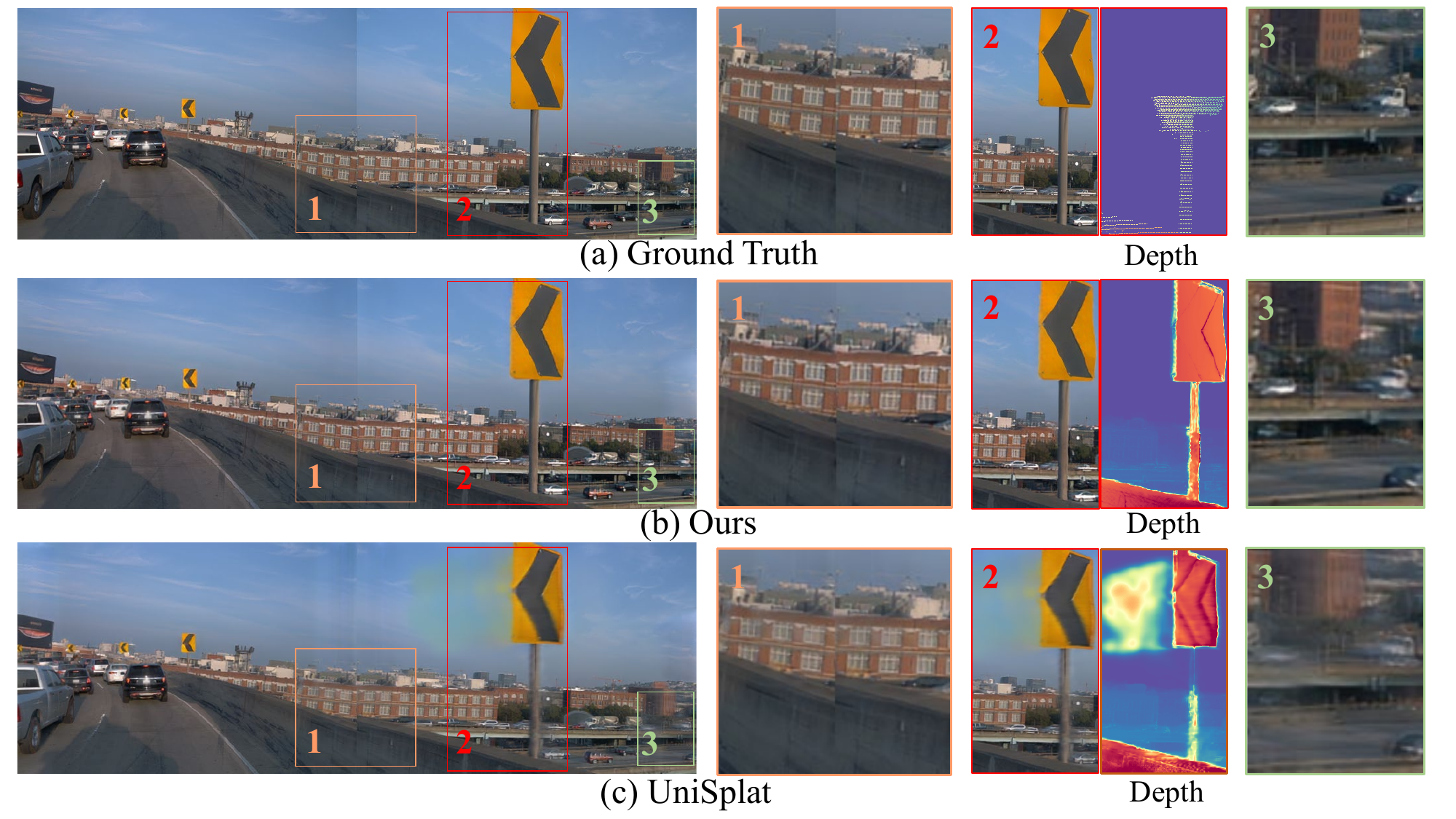}
\vspace{-0.8em}
\caption{
Qualitative comparisons on the Waymo dataset. Compared with existing methods, MCSplat produces finer and cross-camera-consistent results, avoiding texture-induced depth ambiguity within the yellow road sign.
}
\label{fig:visualization1}
\end{figure}

\begin{figure}
\centering
\includegraphics[width=\linewidth]
{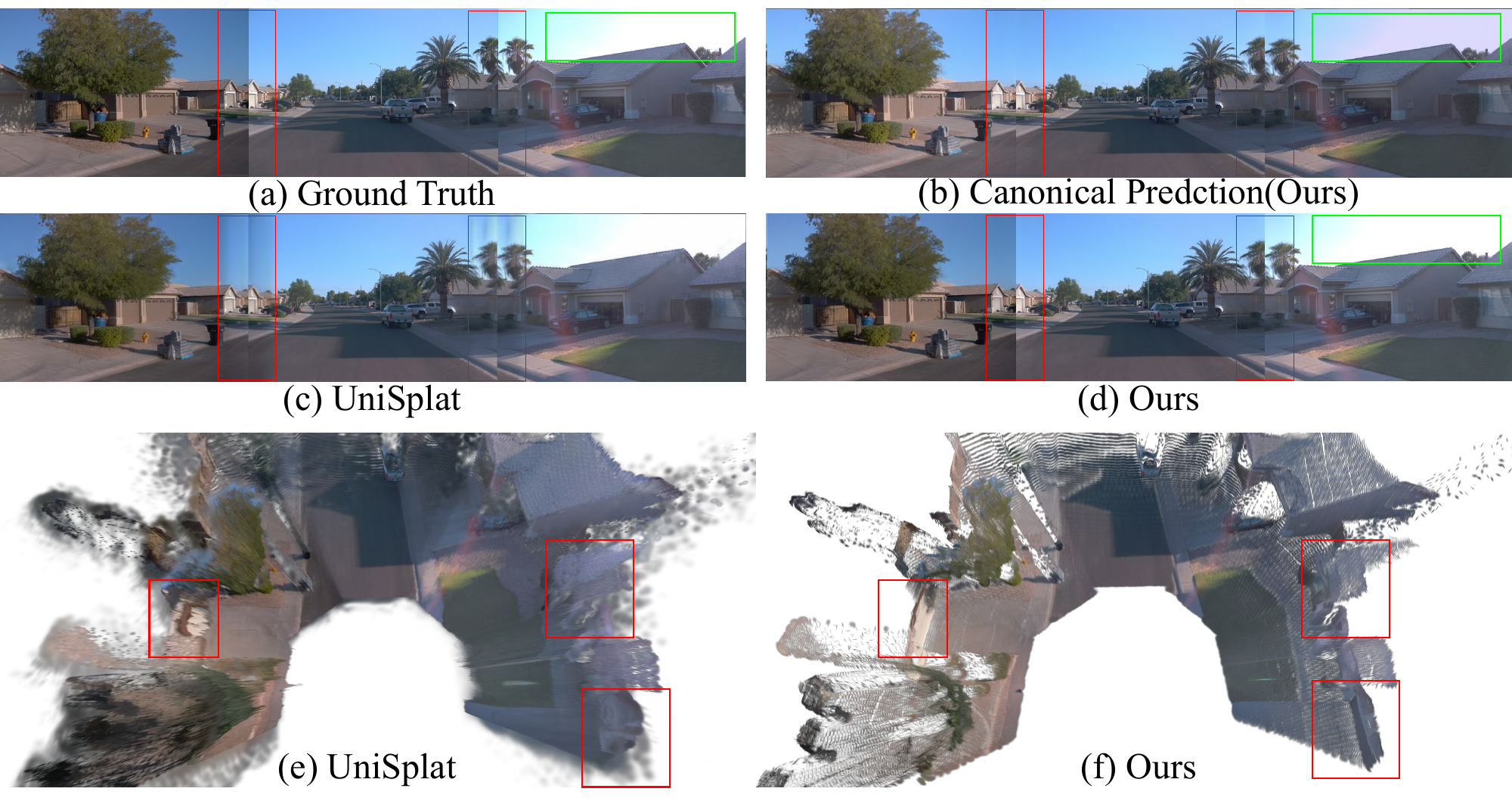}
\caption{
Qualitative results under cross-camera appearance inconsistency. ~\textbf{Zoom in for best view.} Compared with UniSplat (c), our canonical rendering removes camera-boundary artifacts (b), while achieving effective appearance recovery (d).
}
\vspace{-1.0em}
\label{fig:visualization2}
\end{figure}

\paragraph{Image quality on Waymo.}
We compare MCSplat with MVSplat~\citep{chen2024mvsplat}, DepthSplat~\citep{xu2025depthsplat}, EvolSplat~\citep{miao2025evolsplat}, DriveRecon~\citep{lu2024drivingrecon}, and UniSplat~\citep{shi2026unisplat}, using the Waymo results reported by UniSplat for comparison.
As shown in \cref{tab:waymo_nuscenes}, \textbf{under the front-camera setting}, MCSplat improves all reconstruction metrics over UniSplat (PSNR $+$2.12 dB and LPIPS $-33.3\%$) as well as all novel-view synthesis metrics. Although the front-camera input is unchanged (\cref{sec:robust_canonical_appearance}), canonicalizing its neighboring views also benefits reference-view reconstruction. \textbf{Under the 5-camera setting}, MCSplat improves all reconstruction metrics.
Qualitative comparisons in \cref{fig:visualization1,fig:visualization2} illustrate fewer texture-induced geometric artifacts and improved cross-camera consistency; additional visualizations, including cross-lane NVS, are provided in \cref{app:2d_vis,app:3d_vis,app:crosslane_nvs}.

\paragraph{Image quality on nuScenes.}
We additionally evaluate on nuScenes following the protocol used by UniSplat~\citep{shi2026unisplat}: the middle frame of each temporal interval is used as input, and the first, middle, and last frames are evaluation targets. Thus, the nuScenes metrics in \cref{tab:waymo_nuscenes} jointly measure input-view reconstruction and temporal novel-view synthesis. MCSplat achieves 26.19~dB PSNR(+0.82~dB) and 0.783 SSIM(+0.018). This supports generalization beyond the Waymo 5-camera configuration.

\paragraph{Cross-View Geometry and Normal Consistency on Waymo.}
Image-quality metrics alone cannot directly assess Gaussian geometry.
We therefore evaluate cross-view depth reprojection errors and
dense pseudo-normal consistency in \cref{tab:waymo_geometry_normals}.
Since LiDAR observations are too sparse for reliable dense normal estimation,
we use MoGe-2~\citep{wang2025moge2} predictions as pseudo-normal references.
Depth reprojection errors are evaluated in valid overlapping regions
of adjacent cameras.
Compared with UniSplat, MCSplat reduces the mean depth reprojection
error by 61.73\% (2.378 to 0.910~m) and RMSE by 39.01\%
(6.427 to 3.920~m).
Meanwhile, the mean pseudo-normal angular error decreases from
$115.05^\circ$ to $92.82^\circ$, while the fraction of pixels
with angular error below $30^\circ$ increases from 1.18\% to 17.38\%.
These results indicate improved cross-view geometric consistency
and local surface orientation.

\begin{table}[t]
\centering

\begin{minipage}[t]{0.66\linewidth}
\vspace{0pt}
\centering
\caption{
Geometry-oriented evaluation on Waymo.
Normal consistency is measured against
MoGe-2\citep{wang2025moge2} pseudo-normals.
}
\label{tab:waymo_geometry_normals}

\setlength{\tabcolsep}{3pt}
\resizebox{\linewidth}{!}{%
\begin{tabular}{l cc cccc}
\toprule
\multirow{2}{*}{Method} &
\multicolumn{2}{c}{Depth Reproj. (m)} &
\multicolumn{4}{c}{Pseudo-Normal Consistency} \\
\cmidrule(lr){2-3}
\cmidrule(lr){4-7}
& Mean$\downarrow$ & RMSE$\downarrow$ &
Mean ($^\circ$)$\downarrow$ &
$<5^\circ$ (\%)$\uparrow$ &
$<11.25^\circ$ (\%)$\uparrow$ &
$<30^\circ$ (\%)$\uparrow$ \\
\midrule

UniSplat
& 2.378 & 6.427
& 115.05 & 0.02 & 0.10 & 1.18 \\

\rowcolor{BestBG}
\textbf{MCSplat}
& \textbf{0.910} & \textbf{3.920}
& \textbf{92.82}
& \textbf{4.24} & \textbf{10.08} & \textbf{17.38} \\

\bottomrule
\end{tabular}%
}
\end{minipage}
\hfill
\begin{minipage}[t]{0.32\linewidth}
\vspace{0pt}
\centering
\caption{
Ablation on the Stage-1 training objective.
}
\label{tab:appearance_subset}

\setlength{\tabcolsep}{3pt}
\resizebox{\linewidth}{!}{%
\begin{tabular}{lccc}
\toprule
Stage-1 Obj. & PSNR$\uparrow$ &
SSIM$\uparrow$ & LPIPS$\downarrow$ \\
\midrule

Raw GT
& 29.53 & 0.85 & 0.19 \\

Canonical
& \textbf{30.53}
& \textbf{0.88}
& \textbf{0.13} \\

\bottomrule
\end{tabular}%
}
\end{minipage}

\end{table}

\paragraph{Waymo Multi-View Appearance-Inconsistent Subset.}
To assess robustness under strong cross-camera photometric variations, we additionally evaluate MCSplat on a challenging subset of 57 Waymo validation scenes, selected using model-independent cross-camera appearance discrepancy scores (see \cref{app:choose subset}). The detailed reconstruction results are reported in the appendix (\cref{tab:waymo_appearance_subset}).

\subsection{Ablation Study}
We evaluate the roles of Canonical Appearance Construction (CAC), uncertainty-aware geometric self-supervision (Geo), and the Bilateral Grid Transformer (BGT) on Waymo using input-view reconstruction and geometric consistency.

\paragraph{Canonical-Space Training in Stage 1.}
\cref{tab:appearance_subset} compares raw-image versus canonical-image supervision for Stage 1 while keeping the remainder of the training protocol unchanged. Canonical supervision improves PSNR from 29.53 to 30.53~dB, SSIM from 0.85 to 0.88, and LPIPS from 0.19 to 0.13. Training Stage 1 directly on raw, camera-dependent images can introduce conflicting photometric signals in overlap regions and an inconsistent training target for subsequent canonical-space appearance recovery.


\paragraph{Component-Wise Ablation.}
\cref{tab:geo_sem_ablation} evaluates the contributions of CAC, Geo, and BGT to cross-view depth consistency, pseudo-normal
consistency, and input-view reconstruction.
CAC reduces the mean depth reprojection error by 59.34\% (2.378 to 0.967~m) and RMSE from 6.427 to 4.051~m, indicating improved shared geometry through consistent appearance supervision. 
It also reduces the mean pseudo-normal angular error from $115.05^\circ$ to $97.25^\circ$.
Adding Geo further lowers the depth errors to 0.910~m(mean) and 3.920~m (RMSE), while increasing the fraction of pixels with normal error below $30^\circ$ from 4.08\% to 17.38\%.
BGT primarily restores camera-dependent appearance: with CAC and Geo fixed, it improves raw-target PSNR from 29.35 to 30.53~dB, SSIM from 0.86 to 0.88, and LPIPS from 0.14 to 0.13, without changing depth or normal consistency.
Without Geo, CAC+BGT achieves a slightly higher
raw-target PSNR (30.76~dB), suggesting a
geometry--photometry trade-off. These results support the complementary roles of CAC (cross-view geometry), Geo (local surface geometry),
and BGT (appearance recovery).

\begin{table}[t]
\centering
\caption{
Component ablation on the Waymo dataset.
``Can.'' and ``Raw'' denote canonicalized and raw GT images used for evaluation, respectively. Depth reprojection error is evaluated in adjacent-camera
overlap regions, while normal consistency is measured against MoGe-2~\citep{wang2025moge2} pseudo-normals.
}
\label{tab:geo_sem_ablation}
\setlength{\tabcolsep}{3pt}
\resizebox{\linewidth}{!}{%
\begin{tabular}{l c cc ccc cccc}
\toprule

\multirow{2}{*}{Method} &
\multirow{2}{*}{Eval.} &
\multicolumn{2}{c}{Depth Reproj. (m)} &
\multicolumn{3}{c}{Reconstruction} &
\multicolumn{4}{c}{Pseudo-Normal Consistency} \\

\cmidrule(lr){3-4}
\cmidrule(lr){5-7}
\cmidrule(lr){8-11}

& &
Mean$\downarrow$ & RMSE$\downarrow$ &
PSNR$\uparrow$ & SSIM$\uparrow$ & LPIPS$\downarrow$ &
Mean ($^\circ$)$\downarrow$ &
$<5^\circ$ (\%)$\uparrow$ &
$<11.25^\circ$ (\%)$\uparrow$ &
$<30^\circ$ (\%)$\uparrow$ \\

\midrule

Baseline & Raw
& 2.378 & 6.427
& 29.58 & 0.86 & 0.17
& 115.05 & 0.02 & 0.10 & 1.18 \\

\midrule

Baseline + CAC & Can.
& \textbf{0.967} & \textbf{4.051}
& \textbf{31.06} & \textbf{0.88} & \textbf{0.13}
& \textbf{97.25} & \textbf{0.17} & \textbf{0.74} & \textbf{4.08} \\

Baseline + CAC + BGT & Raw
& 0.967 & 4.051
& 30.76 & \textbf{0.88} & \textbf{0.13}
& 97.25 & 0.17 & 0.74 & 4.08 \\

\midrule

Baseline + CAC + Geo & Can.
& 0.910 & 3.920
& 31.09 & 0.88 & 0.14
& 92.82 & 4.24
& 10.08 & 17.38 \\

Baseline + CAC + Geo & Raw
& 0.910 & 3.920
& 29.35 & 0.86 & 0.14
& 92.82 & 4.24
& 10.08 & 17.38 \\

\rowcolor{BestBG}
\textbf{MCSplat (CAC + Geo + BGT)} & Raw
& \textbf{0.910} & \textbf{3.920}
& \textbf{30.53} & \textbf{0.88} & \textbf{0.13}
& \textbf{92.82} & \textbf{4.24}
& \textbf{10.08} & \textbf{17.38} \\

\bottomrule
\end{tabular}%
}
\end{table}

\section{Conclusion}

We presented \textbf{MCSplat}, a multi-view consistent feed-forward Gaussian splatting framework for driving-scene reconstruction. MCSplat decouples geometry learning from camera-specific appearance by constructing a canonical appearance space, recovering raw camera appearance with a multi-scale Bilateral Grid Transformer, and applying Fisher-information guided geometric self-supervision to uncertain Gaussians. Experiments on Waymo and nuScenes demonstrate improved reconstruction quality and cross-dataset generalization, while geometry-oriented evaluations show enhanced cross-view depth consistency and local surface-normal agreement. 
Future work will explore stronger temporal appearance modeling and explicit modeling of ISP parameters.

\begin{ack}
This work was supported by the Fundamental and Interdisciplinary
Disciplines Breakthrough Plan of the Ministry of Education of China
(Grant No. JYB2025XDXM123),
the Science and Technology Commission of Shanghai Municipality
(Grant No. 25ZR1401356),
the Shanghai Municipal Science and Shanghai Automotive Industry
Science and Technology Development Foundation
(Grant No. 2601),
the National Key R\&D Program of China
(Grant No. 2024YFE0115400),
the National Natural Science Foundation of China
(Grant No. 62273256),
and the Jiangxi Provincial Major Science and Technology
Special Project (Grant No. 20262ABE010004).
We are particularly grateful to Zaofu Intelligent Technology Co., Ltd.
for its valuable support.
\end{ack}

\bibliographystyle{unsrt}
{\small
\bibliography{references.bib}
}

\pagebreak
\appendix

\section{Technical appendices and supplementary material}
This supplementary material provides additional technical details, implementation clarifications, and qualitative results that complement the main paper.

\cref{app:compatibility} discusses the compatibility of MCSplat with existing GFM-based feed-forward driving Gaussian frameworks.

\cref{app:fisher_reweighting} explains how Fisher information is computed and used for within-step geometric reweighting.

\cref{app:training_objective} provides detailed definitions of the training loss terms.

\cref{app:implementation_details} provides implementation details, including model configuration, training stages, optimization settings, and hyperparameters.

\cref{app:choose subset} describes the construction and quantitative evaluation of the appearance-challenging validation subset for cross-camera photometric inconsistency.

\cref{app:2d_vis} presents additional 2D rendering results on the Waymo validation set.

\cref{app:3d_vis} presents additional 3D Gaussian visualizations under novel views.

\cref{app:crosslane_nvs} presents qualitative cross-lane novel-view synthesis under laterally shifted camera trajectories.

\subsection{Compatibility with GFM-Based Feed-Forward Driving Gaussian Frameworks}
\label{app:compatibility}

\begin{figure}[!b]
\centering
\includegraphics[width=\linewidth]{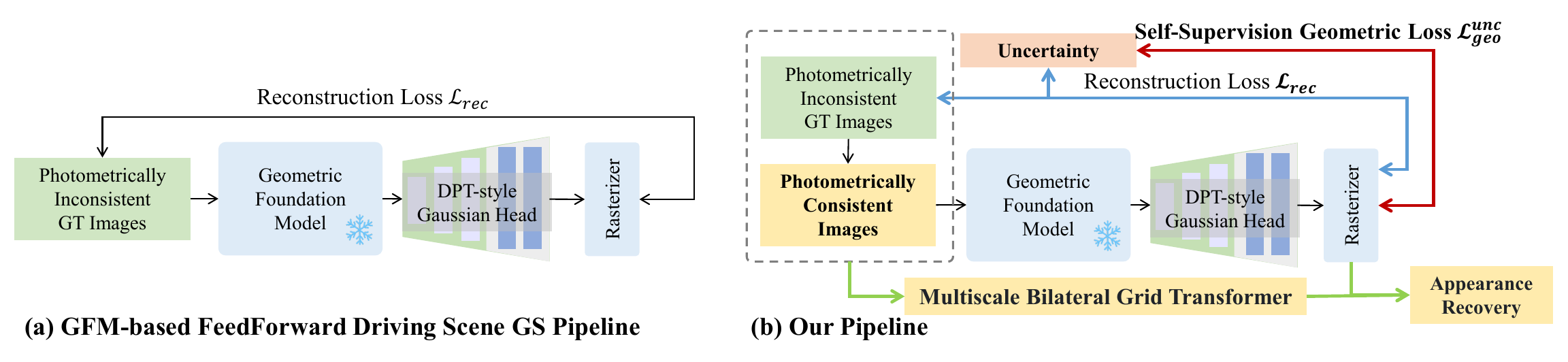}
\caption{
Intuitive pipeline comparison between our method and existing GFM-based feed-forward driving Gaussian frameworks.
}
\label{fig:compatibility}
\end{figure}

Existing GFM-based feed-forward driving Gaussian frameworks\cite{chen2025dggt,shi2026unisplat,yu2026streetforward,yu2026recondrive} usually follow a common pipeline: multi-view images are first encoded by a pretrained geometry foundation model, then decoded by a dense prediction or DPT-style Gaussian head to produce image-aligned Gaussian primitives, which are finally supervised through differentiable rendering and reconstruction losses. Although different methods may vary in temporal fusion, Gaussian parameterization, or auxiliary dynamic modules, their core Gaussian prediction process can be abstracted as an image-to-Gaussian feed-forward mapping followed by rasterization, as shown in \cref{fig:compatibility}(a).

\textbf{MCSplat is designed to be compatible with this general pipeline.} As shown in \cref{fig:compatibility}(b), we keep the GFM backbone and Gaussian prediction branch unchanged, and introduce three modular additions: canonical appearance construction provides photometrically consistent inputs and supervision, the multi-scale Bilateral Grid Transformer maps canonical renderings back to raw camera appearance, and Fisher-information-based uncertainty reweights geometric self-supervision. Therefore, MCSplat can enhance existing GFM-based feed-forward Gaussian frameworks by decoupling geometry learning from camera-specific appearance variation while preserving their original feed-forward reconstruction structure. The ablation results in \cref{tab:geo_sem_ablation} further validate the effectiveness of the proposed modules.

\subsection{Within-Step Fisher Information-Guided Reweighting}
\label{app:fisher_reweighting}
A practical question is how Fisher information computed at the current training step can be used to weight geometric regularizers in the same step. Our implementation adopts a single-step online reweighting strategy rather than a recursive optimization. At iteration \(t\), a standard forward pass first produces the current Gaussian set \(\mathcal{G}^{(t)}\), rendered images \(\hat{I}^{(t)}\), and raw per-Gaussian geometric penalties \(\ell_{i,\mathrm{orient}}^{(t)}\) and \(\ell_{i,\mathrm{flat}}^{(t)}\). Before the main backward pass, we query the gradient of the photometric loss with respect to Gaussian centers on the same forward graph to obtain Fisher information.

Specifically, we use the translational Fisher information as the uncertainty cue:
\begin{equation}
F_i^{(t),xyz}
=
\frac{1}{|\mathcal{V}_{\mathrm{cur}}|}
\sum_{q \in \mathcal{V}_{\mathrm{cur}}}
\left\|
\frac{\partial \mathcal{L}_{\mathrm{photo},q}^{(t)}}{\partial \boldsymbol{\mu}_i^{(t)}}
\right\|_2^2,
\end{equation}
where \(\boldsymbol{\mu}_i^{(t)}\) is the center of Gaussian \(g_i\) at the current step. We then convert the Fisher information into an uncertainty weight:
\begin{equation}
w_i^{(t)}
=
\operatorname{stopgrad}\!\left[
\exp\!\left(
-\gamma
\frac{F_i^{(t),xyz}}{\max_j F_j^{(t),xyz} + \varepsilon}
\right)
\right],
\end{equation}
where \(\operatorname{stopgrad}(\cdot)\) denotes the stop-gradient operator. This detachment is essential: the Fisher information serves only as a current-step importance weight and does not introduce higher-order gradients through the computation.

The orientation and flatness losses are then obtained by weighting the raw per-Gaussian penalties from the same forward pass:
\begin{equation}
\mathcal{L}_{\mathrm{orient}}^{(t)}
=
\frac{\sum_i w_i^{(t)} \, \ell_{i,\mathrm{orient}}^{(t)}}{\sum_i w_i^{(t)}+\varepsilon},
\qquad
\mathcal{L}_{\mathrm{flat}}^{(t)}
=
\frac{\sum_i w_i^{(t)} \, \ell_{i,\mathrm{flat}}^{(t)}}{\sum_i w_i^{(t)}+\varepsilon}.
\end{equation}
Thus, the current-step Fisher information does not recursively define the geometric loss. Instead, it provides an online estimation of geometric uncertainty that reweights the geometric regularizers before the final loss is backpropagated:
\begin{equation}
\mathcal{L}^{(t)}
=
\mathcal{L}_{\mathrm{base}}^{(t)}
+
\lambda_{\mathrm{orient}} \mathcal{L}_{\mathrm{orient}}^{(t)}
+
\lambda_{\mathrm{flat}} \mathcal{L}_{\mathrm{flat}}^{(t)},
\end{equation}
where \(\mathcal{L}_{\mathrm{base}}^{(t)}\) denotes the photometric and auxiliary losses used in the corresponding training stage. This design keeps the optimization first-order while allowing geometric self-supervision to adapt to the local uncertainty of the current reconstruction state.

\subsection{Full Stage-wise Training Objective}
\label{app:training_objective}

Here we present the extended training objectives used in the two-stage optimization. We denote the current-frame and future-frame view sets as \(\mathcal{V}_{\mathrm{cur}}\) and \(\mathcal{V}_{\mathrm{fut}}\), respectively.

\paragraph{Stage 1.}
In the first stage, the GFM backbone is frozen, while the
sparse 3D convolution modules and Gaussian head are
trained in the canonical appearance space. The bilateral-grid branch is bypassed. The objective is defined as
\begin{align}
\mathcal{L}_{\mathrm{stage1}}
={}&
\sum_{v\in\mathcal{V}_{\mathrm{cur}}}
\left(
\lambda_1 \mathcal{L}_{\mathrm{mse}}^{v,\mathrm{can}}
+
\lambda_2 \mathcal{L}_{\mathrm{lpips}}^{v,\mathrm{can}}
+
\lambda_3 \mathcal{L}_{\mathrm{dyn}}^{v}
+
\lambda_4 \mathcal{L}_{\mathrm{scale}}^{v}
\right) \\
&+
\sum_{v\in\mathcal{V}_{\mathrm{fut}}}
\left(
\lambda_1 \mathcal{L}_{\mathrm{mse}}^{v,\mathrm{can}}\odot B^v
+
\lambda_2 \mathcal{L}_{\mathrm{lpips}}^{v,\mathrm{can}}
\right).
\end{align}
Here, \(\odot\) denotes element-wise multiplication, and the background mask \(B^v\) excludes dynamic regions in future-frame supervision to avoid optimization instability.

\paragraph{Stage 2.}
In the second stage, most of the Gaussian head is frozen except for the Gaussian MLPs, while the bilateral-grid transformer is optimized for appearance recovery. The predicted bilateral grids map the canonical render to the raw image domain as in \cref{eq:trans}. The objective is
\begin{align}
\mathcal{L}_{\mathrm{stage2}}
={}&
\mathcal{L}_{\mathrm{stage1}}
+
\sum_{v\in\mathcal{V}_{\mathrm{cur}}\cup\mathcal{V}_{\mathrm{fut}}}
\left(
\lambda_5 \mathcal{L}_{\mathrm{mse}}^{v,\mathrm{img}}
+
\lambda_6 \mathcal{L}_{\mathrm{lpips}}^{v,\mathrm{img}}
\right)
\nonumber\\
&+
\lambda_7 \mathcal{L}_{\mathrm{tv}}
+
\mathcal{L}_{\mathrm{geo}}^{\mathrm{unc}}.
\end{align}
The image-domain losses \(\mathcal{L}_{\mathrm{mse}}^{v,\mathrm{img}}\) and \(\mathcal{L}_{\mathrm{lpips}}^{v,\mathrm{img}}\) supervise the recovered raw-domain image. The total-variation loss \(\mathcal{L}_{\mathrm{tv}}\) regularizes the predicted bilateral grids, and \(\mathcal{L}_{\mathrm{geo}}^{\mathrm{unc}}\) regularizes Gaussian geometry using uncertainty-aware orientation and flatness constraints. \textbf{We detail these loss terms below.}
\paragraph{Details about each loss term.}
Let \(\hat{I}_{\mathrm{can}}^{v}\) and \(I_{\mathrm{can}}^{v}\) denote the canonical prediction and canonical target, and let 
\(\hat{I}_{\mathrm{img}}^{v}=\mathbf{T}_{q,p}(\hat{I}_{\mathrm{can}}^{v})\) be the raw-domain image recovered by the predicted bilateral grids 
\(\mathbf{B}_q^{(m)}\),
with target \(I_{\mathrm{img}}^{v}\). 
Let \(B^{v}(p)\in\{0,1\}\) be the exclusion mask (used only for future-view MSE) for pixel $p$ in the future-view $\mathcal{P}^v$. 
The individual loss terms can be calculated as
\begin{align}
\mathcal{L}_{\mathrm{mse}}^{v,\mathrm{can}}
&=
\sum_{p\in\mathcal{P}^v}
\left\|
\hat{I}_{\mathrm{can}}^{v}(p)
-
I_{\mathrm{can}}^{v}(p)
\right\|_2^2, \\
\mathcal{L}_{\mathrm{mse}}^{v,\mathrm{can}}\odot B^v
&=
\sum_{p\in\mathcal{P}^v}
B^v(p)
\left\|
\hat{I}_{\mathrm{can}}^{v}(p)
-
I_{\mathrm{can}}^{v}(p)
\right\|_2^2, \\
\mathcal{L}_{\mathrm{lpips}}^{v,\mathrm{can}}
&=
\mathrm{LPIPS}\!\left(
\hat{I}_{\mathrm{can}}^{v},
I_{\mathrm{can}}^{v}
\right),\\
\mathcal{L}_{\mathrm{dyn}}^{v}
&=
-\sum_{p\in\mathcal{P}^v}
\Big[
\big(1-B^v(p)\big)\log \hat{D}^{v}(p)
+
B^v(p)\log\big(1-\hat{D}^{v}(p)\big)
\Big],\\
\mathcal{L}_{\mathrm{mse}}^{v,\mathrm{img}}
&=
\sum_{p\in\mathcal{P}^v}
\left\|
\hat{I}_{\mathrm{img}}^{v}(p)
-
I_{\mathrm{img}}^{v}(p)
\right\|_2^2,\\
\mathcal{L}_{\mathrm{lpips}}^{v,\mathrm{img}}
&=
\mathrm{LPIPS}\!\left(
\hat{I}_{\mathrm{img}}^{v},
I_{\mathrm{img}}^{v}
\right).
\end{align}

Since the point clouds estimated by geometry foundation models are often not metrically accurate, we first use the ROE solver from MoGe~\citep{wang2025moge} to compute the optimal scale and shift parameters \((s_v^{*}, t_v^{*})\) between the estimated point cloud and the LiDAR points:
\begin{align}
(s_v^{*}, t_v^{*})
&=
\arg\min_{s,t}
\sum_{p\in\mathcal{P}^{v}}
\frac{1}{Z^{v}(p)}
\left(
s\,\hat{Z}_{\mathrm{local}}^{v}(p)
+
t
-
Z^{v}(p)
\right)^2.
\end{align}
Here, \(\hat{Z}_{\mathrm{local}}^{v}\) denotes the local depth predicted by the Gaussian head before applying the learned scale and shift, and \(Z^{v}\) denotes the corresponding depth supervision. 
A small MLP predicts scale \(\hat{s}^{v}\) and shift \(\hat{t}^{v}\) from the pooled GFM features:
\begin{equation}
(\hat{s}^{v}, \hat{t}^{v})
=
\operatorname{MLP}\!\left(
\operatorname{AvgPool}\!\left(\mathbf{F}_{v}^{\mathrm{geo}}\right)
\right).
\end{equation}

\textbf{The scale-shift loss \(\mathcal{L}_{\mathrm{scale}}^{v}\)} then supervises the predicted alignment parameters to match this closed-form optimum:
\begin{align}
\mathcal{L}_{\mathrm{scale}}^{v}
&=
\mathrm{SmoothL1}\!\left(
\log \hat{s}^{v},
\log s_v^{*}
\right)
+
\mathrm{SmoothL1}\!\left(
\hat{t}^{v},
t_v^{*}
\right).
\end{align}

\textbf{For bilateral-grid smoothness loss \(\mathcal{L}_{\mathrm{tv}}\)}, we regularize the predicted multi-scale bilateral grids with a total-variation penalty. Let \(\mathbf{B}_q^{(m)}\) denote the predicted bilateral grid at scale \(m\), where \(m=1,\dots,M\), and \(q\) indexes the corresponding view or camera-conditioned grid. Each grid stores affine color-transform coefficients over a 3D lattice consisting of the guidance/intensity dimension and two spatial dimensions. We penalize finite differences along all three lattice axes:
\begin{align}
\mathcal{L}_{\mathrm{tv}}
=
\sum_{m=1}^{M}\frac{1}{N_m}
\Bigg(
&
\frac{
\left\|
\mathbf{B}_q^{(m)}{}_{:,:,1:,:,:}
-
\mathbf{B}_q^{(m)}{}_{:,:,:-1,:,:}
\right\|_2^2
}
{
|\nabla_l \mathbf{B}_q^{(m)}|
}
\nonumber
+
\frac{
\left\|
\mathbf{B}_q^{(m)}{}_{:,:,:,1:,:}
-
\mathbf{B}_q^{(m)}{}_{:,:,:,:-1,:}
\right\|_2^2
}
{
|\nabla_h \mathbf{B}_q^{(m)}|
}
\nonumber\\
&+
\frac{
\left\|
\mathbf{B}_q^{(m)}{}_{:,:,:,:,1:}
-
\mathbf{B}_q^{(m)}{}_{:,:,:,:,:-1}
\right\|_2^2
}
{
|\nabla_w \mathbf{B}_q^{(m)}|
}
\Bigg).
\end{align}
Here, \(\nabla_l\), \(\nabla_h\), and \(\nabla_w\) denote finite differences along the guidance, height, and width dimensions of the bilateral grid, respectively; \(|\nabla_l \mathbf{B}_q^{(m)}|\), \(|\nabla_h \mathbf{B}_q^{(m)}|\), and \(|\nabla_w \mathbf{B}_q^{(m)}|\) are the numbers of elements in the corresponding difference tensors, and \(N_m\) is the number of predicted grids at scale \(m\). This loss encourages neighboring lattice cells to predict similar affine color transforms, suppressing high-frequency artifacts in the recovered raw-domain image while still allowing smooth spatial- and intensity-dependent appearance correction.

\textbf{For uncertainty-guided geometric self-supervision loss} \(\mathcal{L}_{\mathrm{geo}}^{\mathrm{unc}}\), please refer to \cref{sec:uncertainty_aware_geo}.

\subsection{Implementation Details}
\label{app:implementation_details}

We provide additional implementation details omitted from the main text. We follow the baseline architecture in the use of the GFM backbone, semantic feature extraction, and Gaussian head design. The input and rendering resolution is fixed to \(350\times518\).

Training is performed in two stages. In Stage 1, the GFM backbone is frozen, and only the 3D sparse convolution modules and the Gaussian head are trained. Different from the baseline, both input and supervision use the canonicalized images from \cref{sec:robust_canonical_appearance}. In Stage 2, all components are frozen except for the Gaussian MLPs and the Bilateral Grid Transformer. This stage jointly learns bilateral-grid appearance recovery and applies geometric regularization to the trainable Gaussian outputs.

The Bilateral Grid Transformer adopts a DINOv2~\citep{oquab2023dinov2} ViT-Small image encoder to extract 384-dimensional patch tokens for appearance conditioning. Both the Transformer encoder and decoder contain 3 layers. The multi-scale bilateral grids use three resolutions \([2,2,1]\), \([4,4,2]\), and \([8,8,4]\), with guidance downsampling factors \([4,4,2]\), respectively.

The model is optimized using AdamW~\citep{loshchilov2017decoupled} on 16 NVIDIA H20 GPUs with a total batch size of 32. For the training objective parameters introduced in~\cref{sec:training_objective} and above, we set
\(\lambda_1=\lambda_5=5.0\),
\(\lambda_2=\lambda_6=0.05\),
\(\lambda_3=0.05\),
\(\lambda_4=0.1\),
\(\lambda_7=1.0\),
\(\lambda_{\mathrm{orient}}=0.05\),
and \(\lambda_{\mathrm{flat}}=0.5\).

\subsection{Validation Subset for Cross-Camera Photometric Inconsistency}
\label{app:choose subset}

\paragraph{Appearance-Inconsistent Subset Selection.}To identify Waymo validation scenes with large cross-camera appearance variations, we compute an overlap-based appearance discrepancy score for each scene. We select 57 scenes using these input-derived, model-independent scores to form a challenging validation subset. Instead of comparing full images directly, we only evaluate pixels that have valid geometric correspondences across cameras. The score is computed separately for ground and sky regions.

For ground regions, we follow the same LiDAR-affine correspondence construction pipeline in \cref{sec:robust_canonical_appearance}. Given a valid LiDAR-depth pixel in a reference camera, we back-project it to 3D and then project it into the target camera. A match is retained only if it lies inside the target image, passes valid-mask filtering, and satisfies the LiDAR depth consistency check when target depth is available. The target RGB value is obtained by bilinear sampling.

For sky regions, where LiDAR depth is unavailable, we use the same proxy sky-point construction as in \cref{sec:robust_canonical_appearance}: each sky pixel is lifted along its camera ray to a fixed sky depth and then projected into the target camera. A correspondence is kept only when both the reference pixel and the projected target pixel are classified as sky.

For each valid correspondence $i$, let the RGB values in the reference and target cameras be
$\mathbf{x}_i=(R_i,G_i,B_i)$ and $\mathbf{y}_i=(R'_i,G'_i,B'_i)$, respectively. We first compute luminance using BT.601 weights:
\[
L(\mathbf{x}_i) = 0.299R_i + 0.587G_i + 0.114B_i,
\]
\[
L(\mathbf{y}_i) = 0.299R'_i + 0.587G'_i + 0.114B'_i.
\]
The per-correspondence appearance difference is defined as the absolute log-luminance difference:
\[
d_i =
\left|
\log_2 \left(\max(L(\mathbf{x}_i), \epsilon)\right)
-
\log_2 \left(\max(L(\mathbf{y}_i), \epsilon)\right)
\right|,
\]
where $\epsilon=10^{-6}$ is used for numerical stability. Equivalently, this measures the absolute logarithmic brightness ratio between two matched observations.

For each scene $s$ and region type $r\in\{\mathrm{ground},\mathrm{sky}\}$, we average over all valid correspondences $\mathcal{M}_{s,r}$:
\[
S_{s,r}
=
\frac{1}{N_{s,r}}
\sum_{i\in \mathcal{M}_{s,r}} d_i,
\qquad
N_{s,r}=|\mathcal{M}_{s,r}|.
\]
The final scene-level score is computed by weighting the ground and sky scores by their numbers of valid correspondences:
\[
S_s
=
\frac{
N_{s,\mathrm{ground}}S_{s,\mathrm{ground}}
+
N_{s,\mathrm{sky}}S_{s,\mathrm{sky}}
}{
N_{s,\mathrm{ground}}+N_{s,\mathrm{sky}}
}.
\]
The final selected Waymo Multi-View Appearance-Inconsistent Subset is listed in \cref{tab:appearance_subset_list}.

\begin{table}[t]
\centering
\caption{
Waymo Multi-View Appearance-Inconsistent Subset.
}
\label{tab:appearance_subset_list}
\scriptsize
\setlength{\tabcolsep}{3pt}
\normalsize
{
\begin{tabular}{rl rl rl}
\toprule
scene\_id & seg\_name & scene\_id & seg\_name & scene\_id & seg\_name \\
\midrule
008 & seg\_107139 & 054 & seg\_143831 & 120 & seg\_283472 \\
009 & seg\_108375 & 059 & seg\_146633 & 122 & seg\_303925 \\
012 & seg\_110487 & 060 & seg\_146873 & 129 & seg\_366934 \\
013 & seg\_110533 & 065 & seg\_150215 & 133 & seg\_419577 \\
014 & seg\_113566 & 067 & seg\_150569 & 134 & seg\_424653 \\
016 & seg\_114061 & 072 & seg\_154962 & 137 & seg\_442641 \\
023 & seg\_121347 & 076 & seg\_159595 & 156 & seg\_599003 \\
025 & seg\_123062 & 078 & seg\_162133 & 157 & seg\_600109 \\
026 & seg\_123746 & 082 & seg\_169798 & 160 & seg\_618300 \\
029 & seg\_128204 & 084 & seg\_171355 & 163 & seg\_662188 \\
033 & seg\_131780 & 088 & seg\_173440 & 164 & seg\_663760 \\
037 & seg\_133368 & 098 & seg\_180241 & 168 & seg\_716314 \\
040 & seg\_134699 & 099 & seg\_180457 & 175 & seg\_798862 \\
041 & seg\_135733 & 109 & seg\_210580 & 180 & seg\_833180 \\
043 & seg\_139416 & 110 & seg\_230820 & 184 & seg\_884527 \\
046 & seg\_140812 & 113 & seg\_250679 & 190 & seg\_902487 \\
047 & seg\_141077 & 115 & seg\_260994 & 195 & seg\_924365 \\
048 & seg\_141279 & 118 & seg\_272435 & 198 & seg\_944394 \\
050 & seg\_142445 & 119 & seg\_273637 & 199 & seg\_947242 \\
\bottomrule
\end{tabular}
}
\end{table}

\paragraph{Reconstruction Results on the Appearance-Inconsistent Subset.}
We compare MCSplat with UniSplat~\citep{shi2026unisplat} on the 57-scene subset. As shown in \cref{tab:waymo_appearance_subset}, MCSplat improves all three reconstruction metrics in both view configurations. \textbf{In the front-camera setting,} PSNR increases from 29.18 to 32.07 dB (+2.89 dB) and LPIPS decreases from 0.15 to 0.12 (-20\%). \textbf{In the 5 camera setting}, PSNR increases from 29.28 to 30.88 dB (+1.60 dB) and LPIPS decreases from 0.18 to 0.14(-22.2\%). The consistent gains on this subset indicate improved robustness to pronounced cross-camera appearance differences.

\begin{table}[!htbp]
\centering
\caption{
Quantitative reconstruction results on the
Waymo Appearance-Inconsistent Subset.
}
\label{tab:waymo_appearance_subset}
\small
\begin{tabular}{l c ccc}
\toprule
Method & Views & PSNR$\uparrow$ & SSIM$\uparrow$ & LPIPS$\downarrow$ \\
\midrule
UniSplat & Front
& 29.18 & 0.89 & 0.15 \\

\rowcolor{BestBG}
Ours & Front
& \textbf{32.07} & \textbf{0.90} & \textbf{0.12} \\

\midrule
UniSplat & Multi
& 29.28 & 0.86 & 0.18 \\

\rowcolor{BestBG}
Ours & Multi
& \textbf{30.88} & \textbf{0.88} & \textbf{0.14} \\

\bottomrule
\end{tabular}
\end{table}
\FloatBarrier

\subsection{Additional Qualitative Results in Image Space}
\label{app:2d_vis}

\begin{figure}[ptb]
\centering
\includegraphics[width=\linewidth]{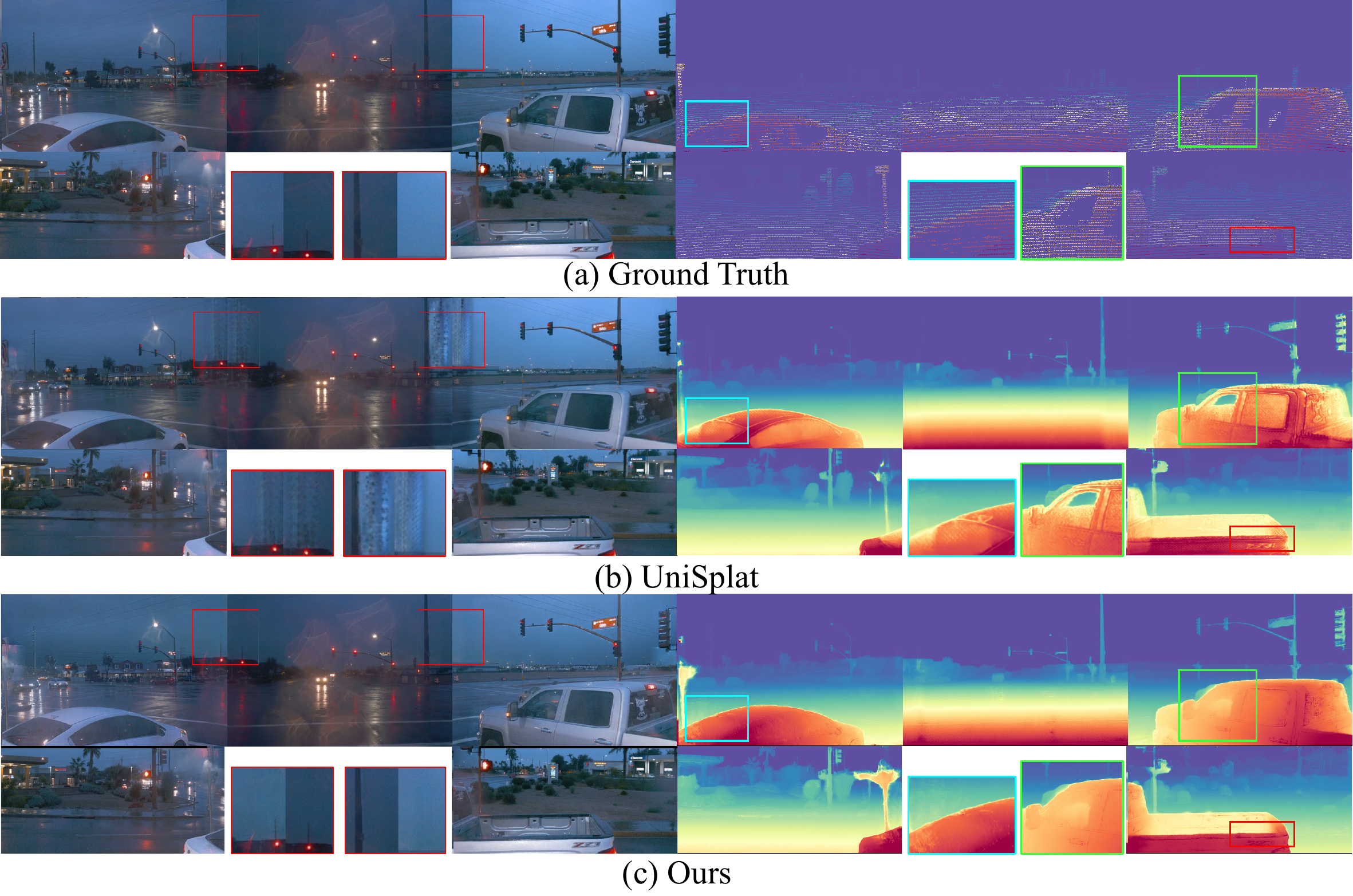}
\caption{
Additional qualitative comparisons under severe cross-camera exposure inconsistency. 
Our method reduces exposure-induced cross-camera artifacts and alleviates texture-induced depth errors. 
Around the pickup truck, it avoids confusing the window depth with the background depth, yielding a more plausible depth estimate.
}
\label{fig:visualization4}
\end{figure}

\begin{figure}[ptb]
\centering
\includegraphics[width=\linewidth]{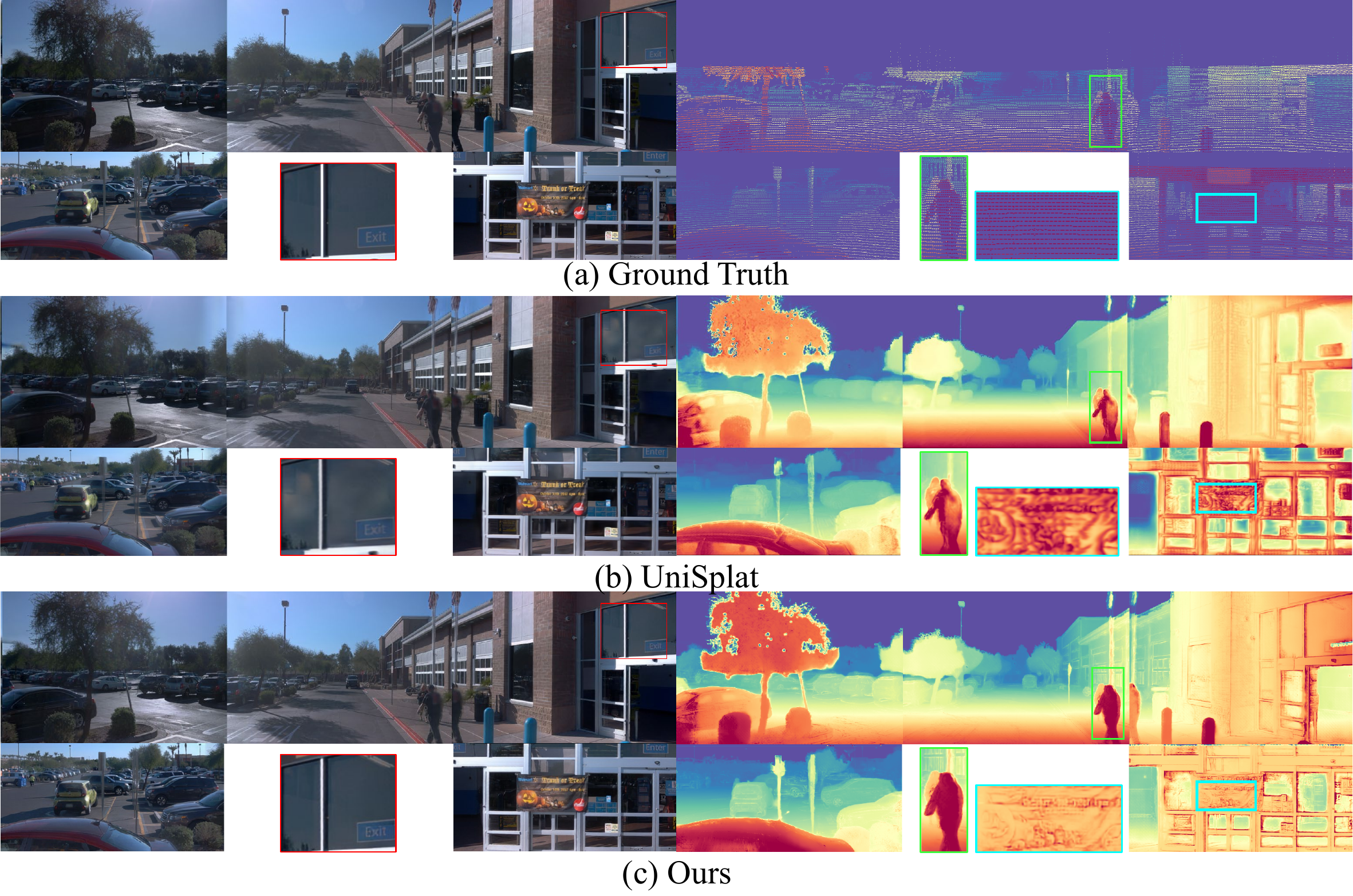}
\caption{
Additional qualitative comparisons.
Our method effectively removes glass-surface artifacts in camera-boundary regions. 
Compared with LiDAR depth, it also yields improved depth estimation in pedestrian regions. 
On the planar poster surface, our method avoids texture-induced depth ambiguity, producing a more reasonable depth prediction.
}
\label{fig:visualization5}
\end{figure}

\begin{figure}[ptb]
\centering
\includegraphics[width=\linewidth]{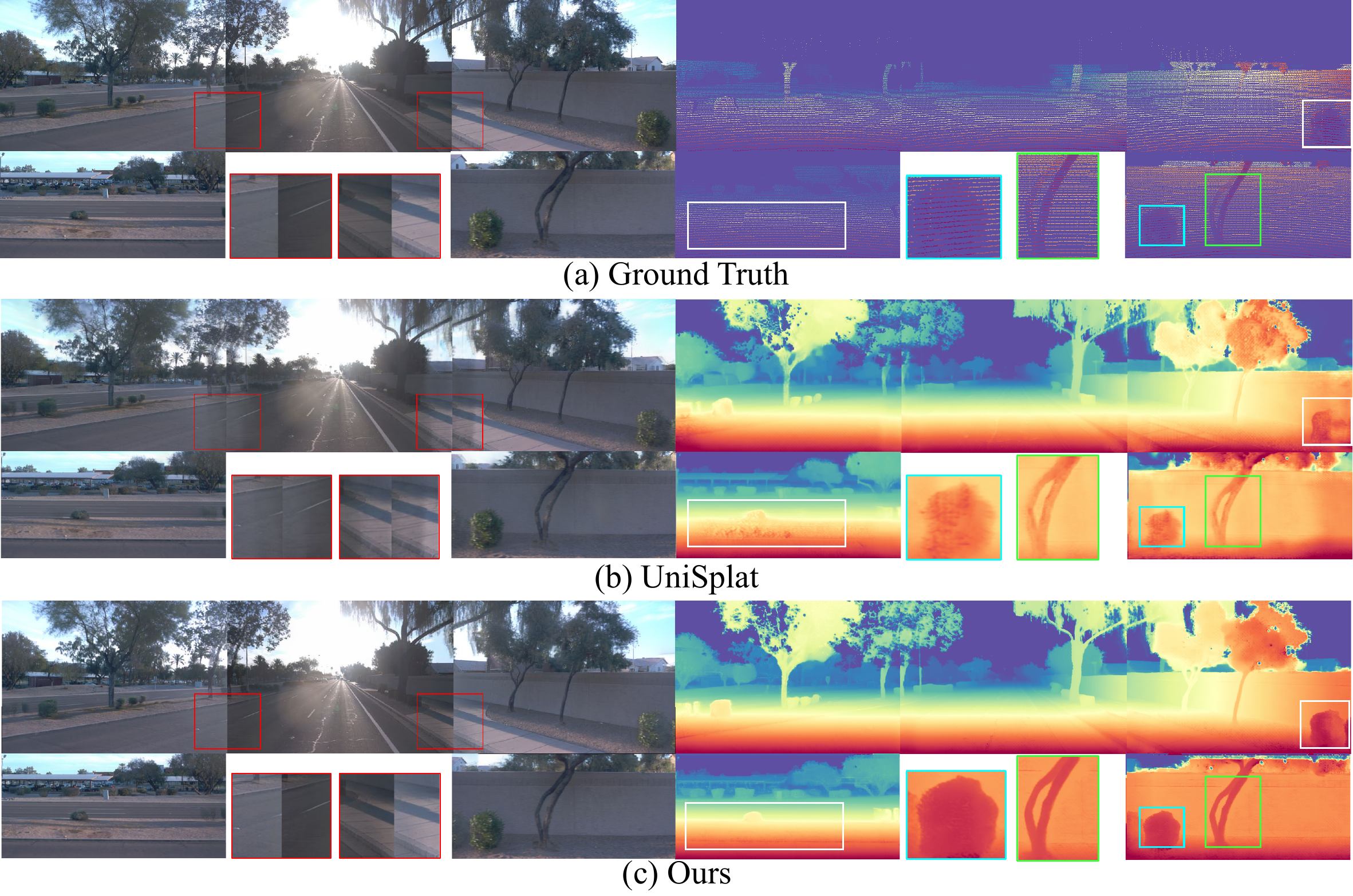}
\caption{
Additional qualitative comparisons on the Waymo dataset. 
Our method reduces cross-camera stitching artifacts and produces more reasonable depth estimates, especially around road boundaries and tree or shrub boundaries.
}
\label{fig:visualization3}
\end{figure}

Additional qualitative results on the Waymo validation set are shown in~\cref{fig:visualization3,fig:visualization4,fig:visualization5}. In the challenging rainy driving scene in~\cref{fig:visualization4}, our method substantially reduces artifacts in the sky compared with the baseline, demonstrating strong appearance recovery capability. Moreover, the introduced uncertainty-guided geometric regularization improves depth estimation around object boundaries, such as shrubs in~\cref{fig:visualization3}, transparent materials such as vehicle windows in~\cref{fig:visualization4}, pedestrians in ~\cref{fig:visualization5}, and texture-rich regions such as the poster in~\cref{fig:visualization5}, which is particularly important for cross-lane novel-view synthesis.

\subsection{Additional Qualitative Results in 3D Gaussian Space}
\label{app:3d_vis}

\begin{figure}[ptb]
\centering
\includegraphics[width=0.7\linewidth]{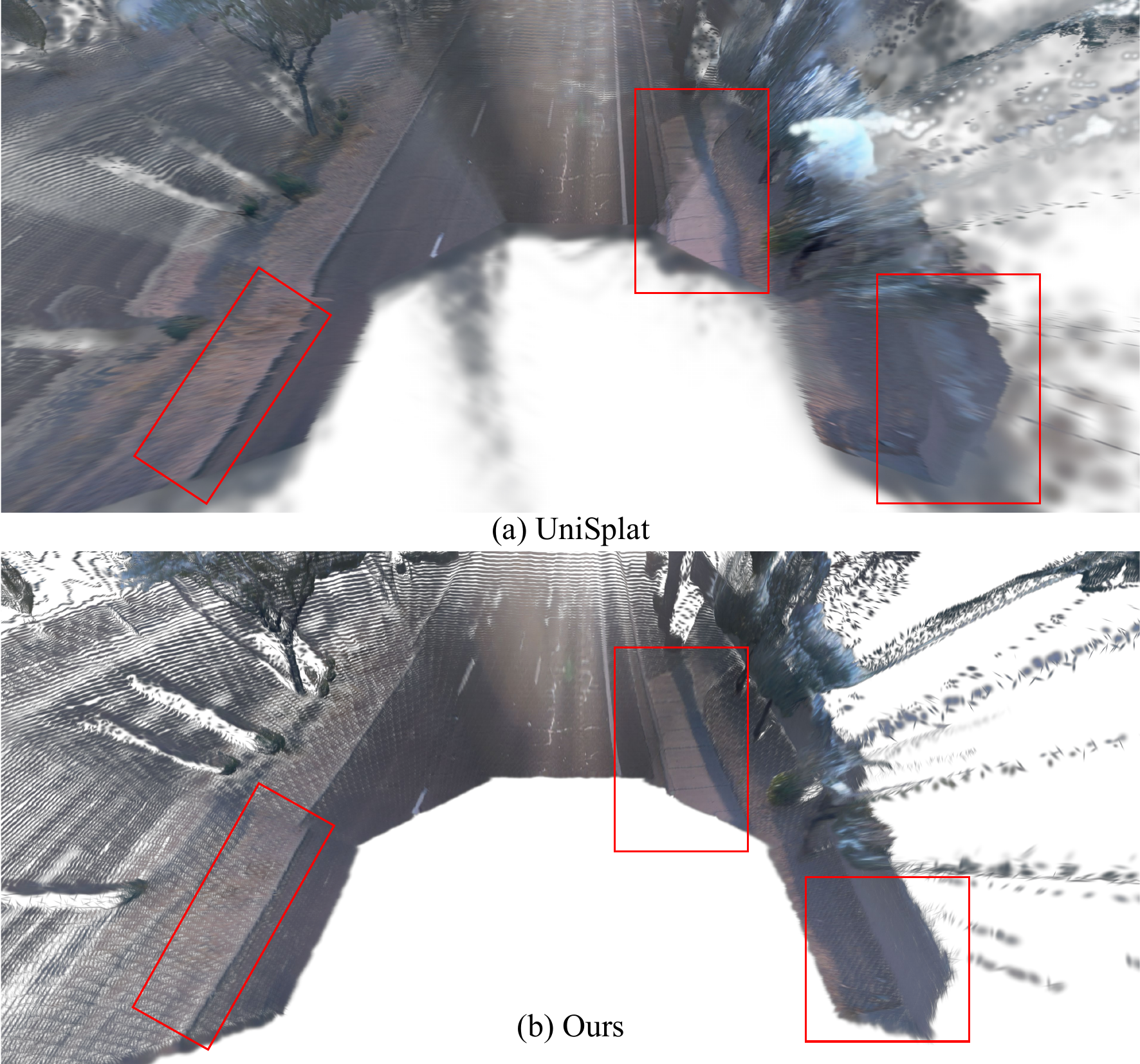}
\caption{
Qualitative comparison of single-frame Gaussians under a novel view. 
Compared with the baseline, our method reconstructs cleaner road boundaries and sharper scene geometry. 
The highlighted regions show that the baseline suffers from geometry--appearance ambiguity caused by multi-view photometric inconsistency, leading to broken road edges, blurred boundary structures, and incomplete ground reconstruction.
}
\label{fig:vis_3d_2}
\end{figure}

\begin{figure}[ptb]
\centering
\includegraphics[width=0.7\linewidth]{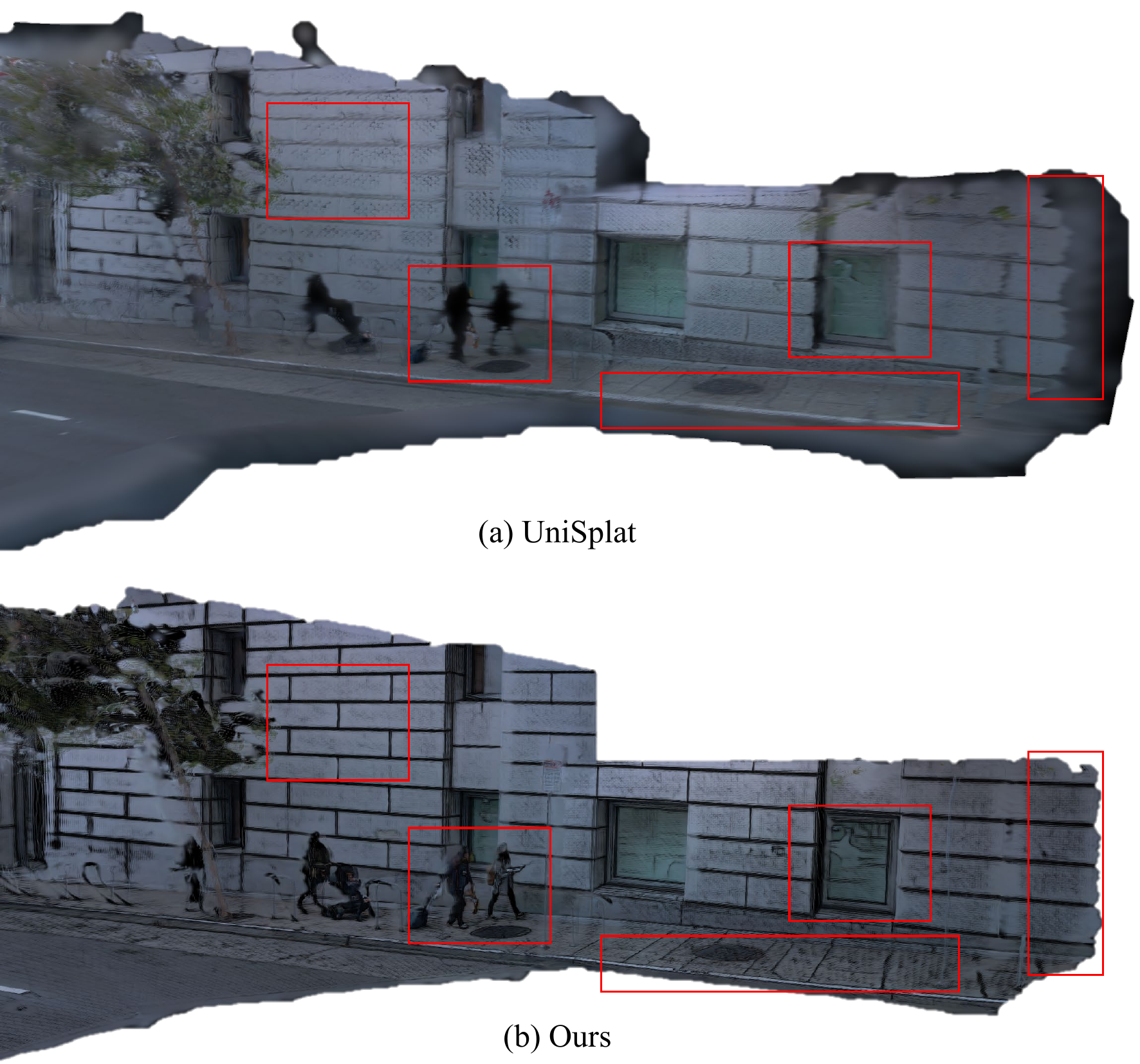}
\caption{
Qualitative comparison of single-frame Gaussians under a novel view in urban scenes.
The highlighted regions show that our method recovers dynamic pedestrians more faithfully than the baseline, while preserving clearer wall textures and more complete ground reconstruction.
}
\label{fig:vis_3d_1}
\end{figure}

\begin{figure}[ptb]
\centering
\includegraphics[width=0.7\linewidth]{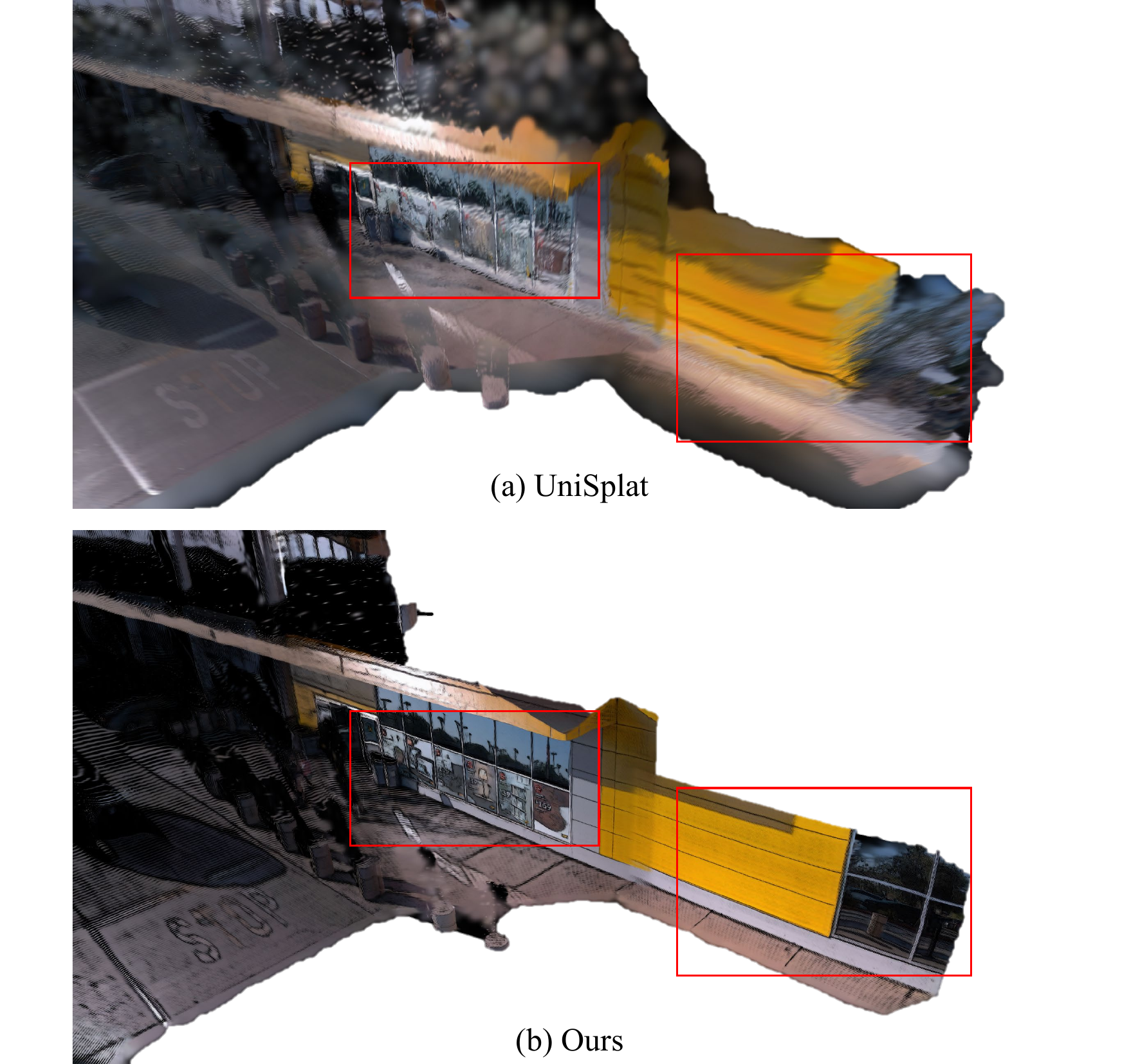}
\caption{
Qualitative comparison of single-frame Gaussians under a novel view. 
UniSplat tends to confuse texture-rich regions with geometric variations, leading to blurred and distorted wall surfaces. 
In contrast, our method preserves sharper texture details and more consistent wall geometry in the highlighted regions.
}
\label{fig:vis_3d_3}
\end{figure}

In~\cref{fig:vis_3d_1,fig:vis_3d_2,fig:vis_3d_3}, we show qualitative comparison of single-frame Gaussians under a novel view in various scenes.
These samples with highlighted regions show that our method recovers the scene more faithfully than the baseline.

\subsection{Cross-Lane Novel-View Synthesis}
\label{app:crosslane_nvs}

\begin{figure}[!htbp]
\centering
\includegraphics[width=0.92\linewidth]{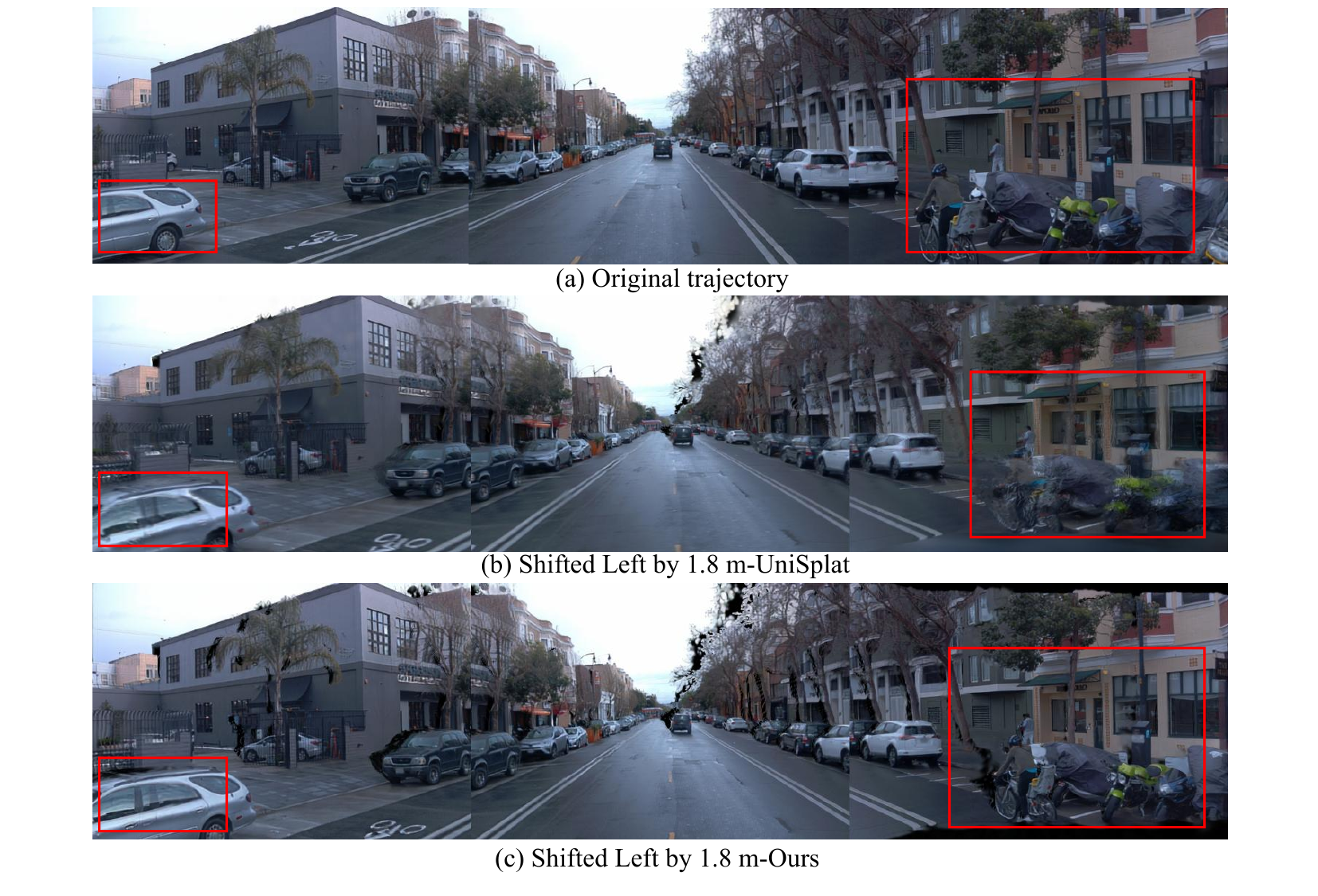}
\caption{
Qualitative cross-lane novel-view synthesis under a 1.8~m leftward camera-trajectory shift. (a) Original trajectory; (b) shifted view rendered by UniSplat; (c) shifted view rendered by MCSplat. The highlighted regions show improved preservation of the left-foreground car and the parked motorcycles along the right sidewalk.
}
\label{fig:crosslane_nvs_1p8}
\end{figure}

\begin{figure}[!htbp]
\centering
\includegraphics[width=0.92\linewidth]{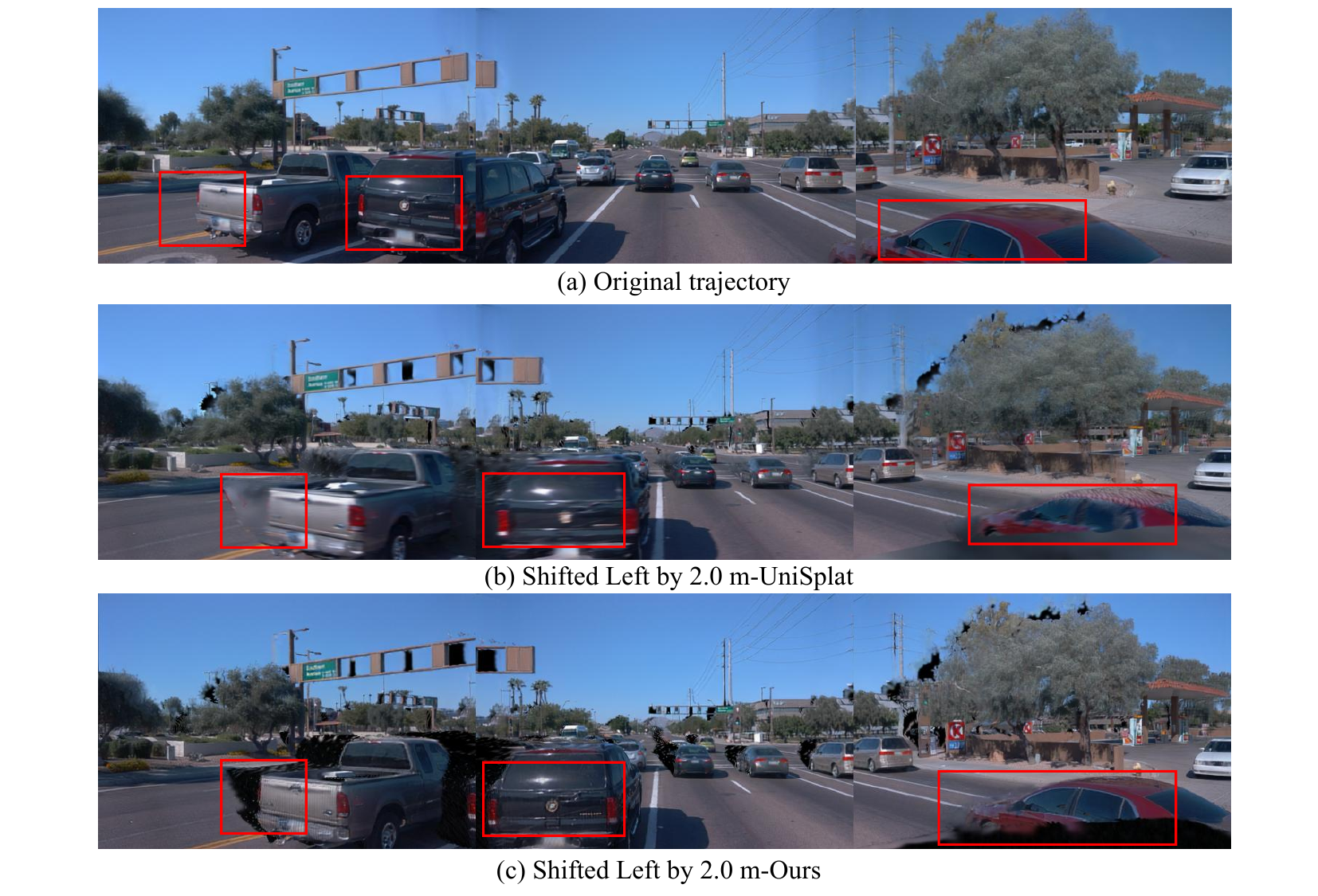}
\caption{
Qualitative cross-lane novel-view synthesis under a 2.0~m leftward camera-trajectory shift. (a) Original trajectory; (b) shifted view rendered by UniSplat; (c) shifted view rendered by MCSplat. Compared with UniSplat, MCSplat better preserves the highlighted pickup-truck rear structures and the red foreground car, although off-trajectory artifacts remain.
}
\label{fig:crosslane_nvs_2p0}
\end{figure}

We evaluate off-trajectory rendering by shifting the camera
1.8~m and 2.0~m leftward
(\cref{fig:crosslane_nvs_1p8,fig:crosslane_nvs_2p0}).
Compared with UniSplat, MCSplat preserves clearer vehicle
contours and fewer geometric distortions, particularly around
the foreground car and curbside motorcycles at 1.8~m,
and the pickup trucks and red car at 2.0~m.
These results suggest improved geometric consistency under
lateral viewpoint shifts, although some artifacts remain
in occluded or sparsely observed regions.




\section{Further Discussion}
\subsection{Social Impact}
\paragraph{Social Impact.}
MCSplat aims to improve feed-forward 3D reconstruction and novel-view synthesis for autonomous driving scenes. Its potential positive impact lies in enabling more reliable closed-loop simulation, safer evaluation of driving policies, and more scalable reconstruction of complex urban environments. By reducing geometry--appearance ambiguity and improving novel-view geometric consistency, the method may help build more faithful simulation assets for testing autonomous systems before real-world deployment.

However, the method should not be viewed as a substitute for real-world safety validation. Reconstruction errors may still occur in rare weather, unusual lighting, long-tail objects, or sensor configurations outside the training distribution. If such reconstructed scenes are used uncritically for safety evaluation, they may lead to overconfidence in downstream autonomous driving systems. In addition, high-fidelity scene reconstruction techniques may raise privacy concerns if applied to data containing identifiable people, vehicles, or locations. Our experiments are conducted on established autonomous driving datasets, and the method is intended for research on reconstruction and simulation rather than surveillance or identification. Responsible deployment should include dataset governance, privacy-preserving data handling, and careful validation across diverse driving conditions.

\subsection{Limitations}
Although MCSplat demonstrates strong cross-camera appearance consistency and camera-specific appearance recovery, its performance can still be affected by the limited appearance diversity of the Waymo dataset. In a few scenes, the recovered raw-domain appearance may be less accurate than expected and exhibit color shifts. A promising direction is to improve appearance coverage through stronger data augmentation and larger, more diverse training datasets.

\FloatBarrier

\newpage
\clearpage

\end{document}